\documentclass[journal]{IEEEtran}
\usepackage{hyperref}       
\usepackage{url}            
\usepackage{booktabs}       
\usepackage{wrapfig}
\usepackage{floatflt}
\usepackage{graphics}
\usepackage{lipsum}
\usepackage{epstopdf}
\usepackage{cite}
\usepackage{color}
\usepackage[mathscr]{eucal}
\usepackage{amsbsy}
\usepackage{bm}
\usepackage{fixltx2e}
\MakeRobust{\overrightarrow}
\usepackage{booktabs}
\usepackage{mathtools}

\usepackage{amssymb}
\usepackage{graphicx}
\usepackage{graphics}
\usepackage{color}
\usepackage{xspace}
\usepackage{bbm}
\usepackage{psfrag}
\usepackage{algorithmicx}
\usepackage{algorithm}
\usepackage{algpseudocode}
\usepackage{multirow}
\usepackage{array}
\usepackage{url}
\usepackage[normalem]{ulem}
\usepackage{floatflt,setspace}
\usepackage{algcompatible}
\usepackage{stackengine}
\def\delequal{\mathrel{\ensurestackMath{\stackon[1pt]{=}{\scriptstyle\Delta}}}}

\algnewcommand\INPUT{\item[\textbf{Input:}]}%
\algnewcommand\OUTPUT{\item[\textbf{Output:}]}%

\usepackage{graphicx}
\usepackage{caption}
\usepackage{subcaption}
\usepackage{amsmath}
\usepackage{amssymb}
\usepackage{array}
\usepackage{booktabs}

\newcommand{\argmin}{{\text{argmin}}}

\usepackage{amsthm}
\newtheorem{definition}{Definition}
\newtheorem{lemma}{Lemma}

\begin{document}

\title{ Tensor Train Neighborhood Preserving Embedding }

\author{Wenqi Wang, Vaneet Aggarwal,  and Shuchin Aeron \thanks{W. Wang and V. Aggarwal are with Purdue University, West Lafayette, IN 47907, email: \{wang2041,vaneet\}@purdue.edu.  S. Aeron is with Tufts
		University, Medford, MA 02155, email: shuchin@ece.tufts.edu. 
		
		The work of W. Wang and V. Aggarwal was supported in part by the U.S. National Science Foundation
under grant CCF-1527486. The work of S. Aeron was supported in part by NSF CAREER Grant \# 1553075 
}}

%

\maketitle

\begin{abstract}
In this paper, we propose a Tensor Train Neighborhood Preserving Embedding (TTNPE) to embed multi-dimensional tensor data into low dimensional tensor subspace. Novel approaches to solve the optimization problem in TTNPE are proposed. For this embedding, we evaluate novel trade-off gain among classification, computation, and dimensionality reduction (storage) for supervised learning.
It is shown that compared to the state-of-the-arts tensor embedding methods, TTNPE achieves superior trade-off in classification, computation, and dimensionality reduction in MNIST handwritten digits and Weizmann face datasets.
\end{abstract}

\section{Introduction}

Robust feature extraction and dimensionality reduction are among the most fundamental problems in machine learning and computer vision. 
Assuming that the data is embedded in a low-dimensional subspace, popular and effective methods for feature extraction and dimensionality reduction are the Principal Component Analysis (PCA) \cite{jolliffe2002principal,bishop2006pattern}, and the Laplacian eigenmaps \cite{belkin2003laplacian}. 
{
However, simply projecting data to a low dimensional subspace may not efficiently extract discriminative features. Motivated by
recent works  \cite{phien2016efficient,wang2016tensor,wang2017tensor} that demonstrate applying tensor
factorization (after reshaping matrices to multidimensional arrays or tensors) improves data
representation, we consider reshaping vision data into tensors and embedding the tensors
into Kronecker structured subspaces, i.e. tensor subspaces, to further refine these subspace
based approaches with significant gains. In this context, a very popular representation
format namely Tucker format has shown to be useful for a variety of applications \cite{de2000multilinear,lu2006multilinear,vasilescu2003multilinear,zeng2014multilinear}. However, Tucker representation is exponential in storage requirements \cite{ashraphijuo2016deterministic}. In  \cite{holtz2012manifolds}, it was shown that hierarchical Tucker representation, and in
particular Tensor Train (TT) representation is a promising format for the approximation of
solutions in high dimensional data and can alleviate the curse of dimensionality under fixed rank, which inspires us to investigate its application in efficient dimensionality reduction
and embedding.  Tensor train representation has also been shown to be useful for dimensionality reduction in \cite{novikov2015tensorizing,tjandra2017compressing,wang2018wide}. 
}

In this paper, we begin by noting that TT decompositions are associated with a structured subspace model, namely the Tensor Train subspace \cite{hackbusch2012tensor}. 
Using this notion, we extend a popular approach, namely the Neighborhood Preserving Embedding (NPE) \cite{he2005neighborhood} for unsupervised classification of data. 
In the past, the NPE approach has been extended to exploit the Tucker subspace structure on the data \cite{he2005tensor,dai2006tensor}. 
Here, we embed the data into a Tensor Train subspace and propose a computationally efficient Tensor Train Neighbor Preserving Embedding (TTNPE) algorithm. 
We show that this approach achieves significant improvement in the storage of embedding and computation complexity for classification after embedding as compared to the embedding based on the Tucker representation in \cite{he2005tensor,dai2006tensor}. 
{ An approximation method for TTNPE, called TTNPE-ATN (TTNPE- Approximated Tensor Networks) is  provided to decrease the computational time for embedding the data. }
We validate the approach on classification of MNIST handwritten digits data set \cite{lecun1998gradient}, Weizmann Facebase \cite{Weizmann}, { and financial market dataset.} 

{The key contributions of this paper are as follows. (i) We formulate the problem of  embedding the data into a low-rank Tensor Train subspace, and propose a TTNPE  algorithm for embedding the data. (ii) We give an approximation method to the embedding algorithm, TTNPE-ATN, to achieve faster computational time. (iii) We show that embedding based on TTNPE-ATN achieves significant improvement in the storage of embedding and computation complexity for classification after embedding as compared to the embedding based on the Tucker representation.  Finally, the results on the different datasets show significant improvement in classification accuracy, computation and storage complexities for a given compression ratio, as compared to the baselines.     }

The rest of the paper is organized as follows. 
The technical notations and definitions are introduced in Section \ref{NP}.  
The  Tensor Train subspace (TT-subspace) is described in Section \ref{TTS}. 
In Section \ref{NPE}, the optimization problem for Tensor Train Neighbor Preserving Embedding (TTNPE) is formulated. We then outline algorithms to solve the resulting problem highlighting the computational challenges and propose an approximate method to alleviate them.
In section \ref{simu}, we evaluate the proposed algorithm on MNIST handwritten digits, Weizmann databases, { and financial market dataset.}
Section \ref{concl} concludes the paper.  

\section{Notations and Preliminaries}\label{NP}
Vectors and matrices are represented by boldface lower letters (e.g. ${\bf x}$) and boldface capital letters (e.g. ${\bf X}$), respectively. 
An $n$-order tensor is denoted by calligraphic letters ${\mathscr{X}} \in \mathbb{R}^{I_1 \times I_2 \times ... \times I_n}$, where $I_{i},\, i=1,2,..., n$ denotes the dimensionality along the $i_\text{th}$ order. 
An element of a tensor $\mathscr{X}$ is represented as $\mathscr{X}(i_1, i_2,\cdots, i_n)$, where $i_{k},\, k=1,2,.., n$ denotes the location index along the $k_{\text{th}}$ order.  A colon is applied to represent all the elements of an order in a tensor,  e.g. $\mathscr{X}(:, i_2,\cdots, i_n)$ represents the fiber along order $1$ and $\mathscr{X}[:, :, i_3, i_4,\cdots, i_n]$ represents the slice along order $1$ and order $2$ and so forth. 
${\bf V}(\cdot)$ is a tensor vectorization operator such that  $\mathscr{X} \in \mathbb{R}^{I_1 \times \cdots \times I_n }$ is mapped { to a} vector ${\bf V}({\mathscr{X}}) \in \mathbb{R}^{I_1  \cdots  I_n }$.
$\times$ and $\otimes$ represent matrix product and kronecker product respectively. 
Let $\text{tr}_{i}^{j}$ be a tensor trace operation, which reduces 2 tensor orders by getting the trace along the slices formed by the $i_\text{th}$ and $j_\text{th}$ order (assuming $I_i=I_j$). As an example,  let $\mathscr{U} \in \mathbb{R}^{I_1 \times I_2 \times I_1}$ be a 3-mode tensor, then ${\bf v} = \text{tr}_1^3 (\mathscr{U}) \in \mathbb{R}^{I_2}$ is given as ${\bf v}(i_2) = \text{trace}(\mathscr{U}(:, i_2, :)), i_2 = 1,\cdots, I_2$.

\begin{figure*}[t!]
\includegraphics [trim=.1in .2in .1in 1in, keepaspectratio, width=0.6\textwidth] {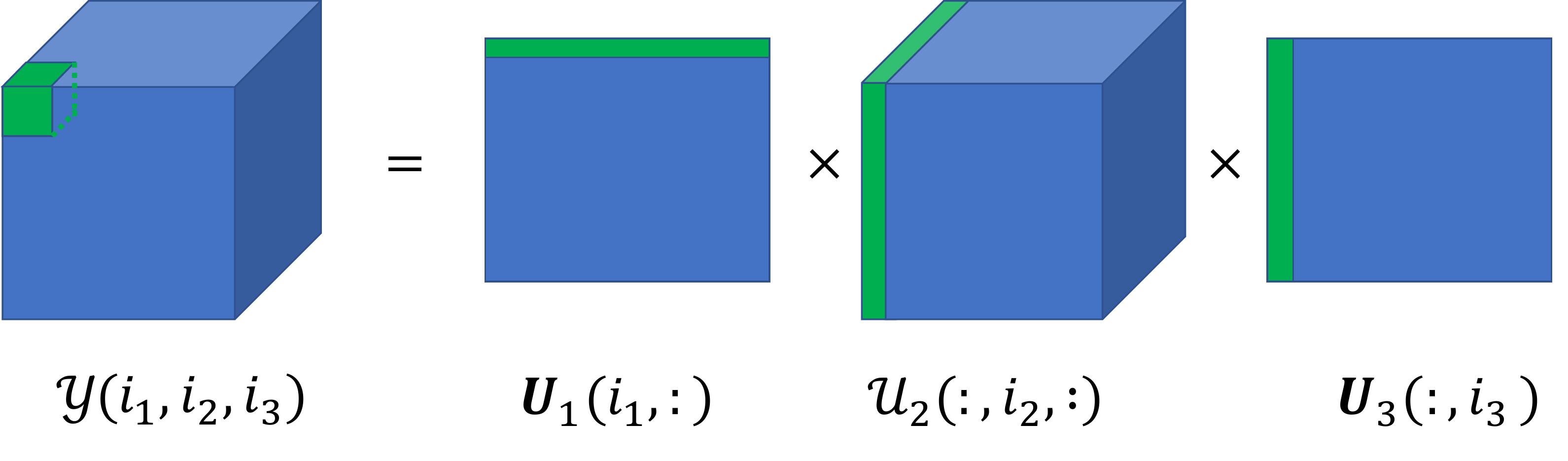}
\centering
\caption{\small {{ Tensor train decomposition for a $3$-mode tensor.
}}}
\label{TT_decom}
\end{figure*}

\begin{figure*}[t!]
\includegraphics [trim=.1in .2in .1in .3in, keepaspectratio, width=0.8\textwidth] {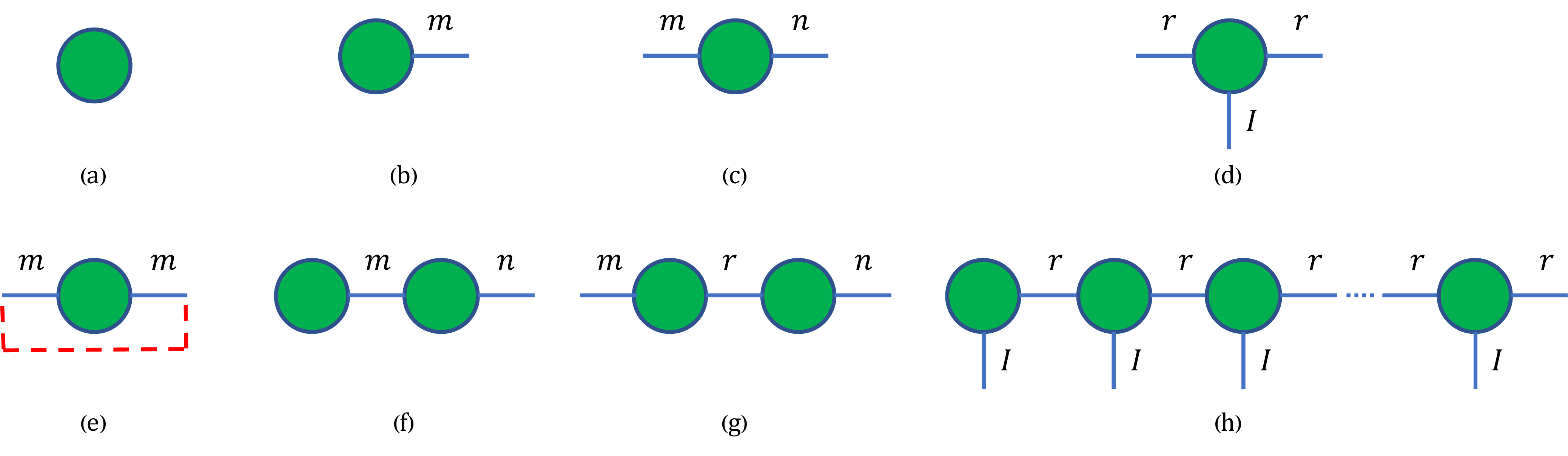}
\centering
\caption{\small {{Tensor network notations. Each node represents a tensor and the number of edges determines the mode of a tensor. The edge connecting two nodes is the operation of tensor merging product. (a) scalar $s\in\mathbb{R}^0$, (b) vector ${\bf v}\in\mathbb{R}^{m}$, (c) matrix ${\bf M} \in \mathbb{R}^{m\times n}$, (d) tensor $\mathscr{T} \in \mathbb{R}^{r \times I \times r}$,  (e) Trace operation $\text{tr}({\bf M})$, (f) vector to matrix product between ${\bf v}\in \mathbb{R}^{m}$ and ${\bf M}\in \mathbb{R}^{m \times n}$, (g) product of two matrices ${\bf M}_1 \in \mathbb{R}^{m\times r}$  and ${\bf M}_2 \in \mathbb{R}^{r\times n}$, (h) tensor merging product for tensor train decomposition $\{{\bf U}_1,\mathscr{U}_2,\cdots, \mathscr{U}_{n-1}, \mathscr{U}_n\}$.}}}
\label{TN}
\end{figure*}

We first introduce the tensor train decomposition.
\begin{definition}\label{TTD} (Tensor Train (TT) Decomposition \cite{oseledets2011tensor,holtz2012manifolds}) Each element of a $n$-mode tensor $\mathscr{Y} \in \mathbb{R}^{I_1\times \cdots \times I_n}$ in tensor train representation is generated by
\begin{equation}
\begin{split}
&\mathscr{Y}(i_1,\cdots, i_n) =  \\
&{\bf U}_1(i_1,:) \mathscr{U}_2(:, i_2, :) \cdots \mathscr{U}_{n-1}(:, i_{n-1}, :){\bf U}_n(:, i_n),
\end{split}
\end{equation}
where ${\bf U}_1\in\mathbb{R}^{I_1 \times R_1}$ and ${\bf U}_n \in \mathbb{R}^{R_{n-1} \times I_n}$ are the boundary matrices and $\mathscr{U}_i \in \mathbb{R}^{R_{i-1} \times I_i \times R_i}, i=2,\cdots, n-1$ are the decomposed tensors. 
\end{definition}
{
Tensor train decomposition for a $3$-mode tensor $\mathscr{Y}$ is illustrated in Fig. \ref{TT_decom}, where $\mathscr{Y}(i_1, i_2. i_3)$ is the sequential product of vector ${\bf U}_1(i_1, :)$, matrix $\mathscr{U}_2(:, i_2, :)$, and vector ${\bf U}_3(:, i_3)$. }

In this paper, 
we consider { a tensor train decomposition for a tensor data set}, which is an
 $n+1$ mode tensor $\mathscr{X} \in \mathbb{R}^{I_1 \times \cdots \times I_n \times R_n}$, where each element is represented as
\begin{eqnarray}
&&\mathscr{X}(i_1,\cdots, i_n, r_n) \nonumber\\
&=& {\bf U}_1(i_1,:) \mathscr{U}_2(:, i_2, :) \cdots \mathscr{U}_{n-1}(:, i_{n-1}, :)\mathscr{U}_n(:, i_n, r_n).
\end{eqnarray}

Without loss of generality, we let $R_0=1$ and define ${\mathscr{U} }_1\in\mathbb{R}^{R_0\times I_1 \times R_1}$ as the tensor representation of ${\bf U}_1$. 
Thus, the tensor train decomposition for $\mathscr{X} \in \mathbb{R}^{I_1 \times \cdots \times I_n \times R_n}$ is
{
\begin{eqnarray}
&&\mathscr{X}(i_1,\cdots, i_n, r_n) \nonumber\\
&=& \mathscr{U}_1(1,i_1,:) \cdots \mathscr{U}_{n-1}(:, i_{n-1}, :)\mathscr{U}_n(:, i_n, r_n).
\end{eqnarray}
}
The TT-Rank of a tensor is denoted by a vector of ranks $(R_1, \cdots, R_{n})$ in the tensor train decomposition. 
{ Left and right unfoldings reshape tensors into matrix, and are defined as follows.}
\begin{definition}(Left and Right Unfolding) Let $\mathscr{X} \in \mathbb{R}^{I_1 \times \cdots \times I_n \times R_n}$ be a $n+1$ mode tensor. 
The left unfolding operation is the matrix obtained by taking the first $n$ mode as row indices and the last mode as column indices such that
${\bf L}(\mathscr{X}) \in \mathbb{R}^{(I_1\cdots I_n) \times R_n}$.
Similarly, the right unfolding operation produces the matrix obtained by taking the $1$st mode as row indices and the remaining $n$ mode as column indices such that
${\bf R}(\mathscr{X}) \in \mathbb{R}^{I_1 \times (I_2 \cdots I_n R_n)}$.
\end{definition}

We further introduce a tensor operation and show the equivalence of tensor operations to matrix product.


\begin{definition}(Tensor Merging Product) 
Tensor merging product is an operation to merge the two tensors along the given sets of mode indices.  
Let $\mathscr{U}_1 \in \mathbb{R}^{I_1 \times \cdots \times I_n}$ and $\mathscr{U}_2 \in \mathbb{R}^{J_1 \times \cdots \times J_m}$ be two tensors. 
 Let ${\bf g}_{i}$, $i\in\{1,2\}$, be a $k-dimensional$ vector such that ${\bf g}_{i}(p) \in \{1,\cdots, n\}, 1\leq p \leq k$ and $I_{{\bf g}_1(p)} =J_{{\bf g}_2(p)}$.
 Then, the tensor merging product is 
\begin{equation}
\mathscr{U}_3 = \mathscr{U}_{1} \times_{{\bf g}_1}^{{\bf g}_2} \mathscr{U}_2  \in \mathbb{R}^{\{\times_{p\notin {\bf g}_1} I_p\}  \times \{\times_{p\notin {\bf g}_2} J_p\}},
\end{equation}
which is a $m+n-2k$ mode tensor, given as
{
\begin{equation}
\begin{split}
&\mathscr{U}_3(i_t \forall t\notin {\bf g}_1, j_q  \forall q\notin {\bf g}_2) \\
= &\sum_{d_1,\cdots, d_k}\mathscr{U}_1(a_1, \cdots, a_n)\mathscr{U}_2(b_1, \cdots, b_m),
\end{split}
\end{equation}
}
where $a_r = i_r$ for $r\notin {\bf g}_1$, $b_r = j_r$ for $r\notin {\bf g}_2$, 
{ $a_{g_1(p)} = d_p$ for $p=1, \cdots, k$, and $b_{g_2(p)} = d_p$ for $p=1, \cdots, k$}. 
\end{definition}

Based on tensor merging product, we note that recovering a tensor from tensor train decomposition is a process of applying tensor merging product on tensor train factorizations. 
{ For a better understanding of tensor train decomposition, we use the tensor network notation given in  \cite{cichocki2016low} to describe tensor merging product and its relation with tensor train decomposition in Fig. \ref{TN}. }
Let  $R_0=1$, $\mathscr{U}_i \in \mathbb{R}^{R_{i-1} \times I_i \times R_i}, i=1,\cdots, n$,  be $n$ $3$-rd order tensors. The recovery of the $n+1$ order tensor is defined as
{
\begin{equation}
\mathscr{U} = \mathscr{U}_1 \times_3^1 \mathscr{U}_2 \times \cdots \times_3^1  \mathscr{U}_n \in \mathbb{R}^{I_1 \times \cdots \times I_n \times R_n}.
\end{equation}
}
The matrix product between ${\bf A} \in \mathbb{R}^{m \times r}$ and ${\bf B} \in \mathbb{R}^{r\times n}$ is equivalent to ${\bf A} \times_{2}^{1} {\bf B}$. This is because if ${\bf C}= {\bf A} \times {\bf B}$, then ${\bf C}(i,j) = \sum_k {\bf A}(i,k) {\bf B}(k,j)$. Similarly, ${\bf A} \times {\bf B} = {\bf B} \times_{1}^{2} {\bf A}$.

{\lemma \label{equivalence} Let $\mathscr{A} \in \mathbb{R}^{M \times R_1 \times \cdots \times R_k}$ and  $\mathscr{B} \in \mathbb{R}^{R_1 \times \cdots \times R_k \times N}$ be two $k+1$ mode tensors, and let ${\bf A} \in \mathbb{R}^{M \times (R_1\cdots R_k)}$ and ${\bf B} \in \mathbb{R}^{(R_1\cdots R_k) \times N}$ be the right and left unfolding of $\mathscr{A}$ and $\mathscr{B}$. 
Tensor merging product,  $\mathscr{A} \times^{1,\cdots,k}_{2,\cdots,k+1} \mathscr{B}$, is the same as ${\bf A} \times {\bf B}$.
\proof{ Proof is given in Appendix \ref{Lemma1} .} \endproof
}

\section{Tensor Train Subspace (TTS)}\label{TTS}
A tensor train subspace, ${\mathscr{S}_\text{TT}} \subseteq \mathbb{R}^{I_1\times I_2 \times \cdots  \times I_n}$, is defined as the span of a $n$-order tensor that is generated by the tensor merging product of a sequence of $3$-order tensors. Specifically,
{
\begin{equation}
{\mathscr{S}_{\text{TT}}} \delequal 
\{ \mathscr{U}_1\times_3^{1} \mathscr{U}_2 \times \cdots\times_3^1  \mathscr{U}_n \times_3^{1} {\bf a} | \forall {{\bf a} \in \mathbb{R}^{R_n }} \}.
\end{equation}
}

For comparison with vector subspace model, tensors can be vectorized into vectors and the tensor train subspace expressed under matrix form gives
{
\begin{equation}
{{\bf S}_{\text{TT}}} = \{ {\bf L}(\mathscr{U}_1\times_3^{1} \mathscr{U}_2 \times \cdots\times_3^1  \mathscr{U}_n ){\bf a} | \forall {{\bf a} \in \mathbb{R}^{R_n }} \}.
\end{equation}}

We note that a tensor train subspace is determined by  $\mathscr{U}_1, \mathscr{U}_2, \cdots, \mathscr{U}_n$, where $\mathscr{U}_i \in \mathbb{R}^{R_{i-1} \times I_i \times R_i}$, $R_0 =1$. When $n=1$, the proposed tensor train subspace reduces to the linear subspace model under matrix case. 

{\lemma \label{LemmaSP} (Subspace Property) ${\bf S_{\text{TT}}} $ is a { $R_n$ dimensional subspace of $\mathbb{R}^{I_1  \cdots  I_n}$ for a given set of {decomposed tensors.}
}, $\{\mathscr{U}_1, \mathscr{U}_2, \cdots, \mathscr{U}_n \}$. }

We next briefly outline some useful properties of the TT decomposition that will be used in this paper. 

{\lemma (Left-Orthogonality Property \cite[Theorem 3.1]{holtz2012manifolds})
For any tensor 	$\mathscr{X} \in \mathbb{R}^{I_1 \times \cdots \times I_n \times R_n}$ of TT-rank ${\mathbf R} = [R_1,\cdots, R_{n-1}]$, the TT decomposition  can be chosen such that  ${\bf L}(\mathscr{U}_i)$ is left-orthogonal for all $i=1, \cdots n$, or ${\bf L}(\mathscr{U}_i)^\top {\bf L}(\mathscr{U}_i) = {\bf I}_{R_i} \in \mathbb{R}^{R_i \times R_i}$.	}

As a consequence of this result we have the following Lemma.

{\lemma (Left-Orthogonality of Tensor Merging Product) \label{Lemma4}
If ${\bf L}(\mathscr{U}_i)$ is left-orthogonal for all $i=1, \cdots, n$, then  ${\bf L}(\mathscr{U}_1 \times_3^1 \cdots \times_3^1 \mathscr{U}_j)$ is left-orthogonal for all $1\le j\le n$. \label{tcplo}}
\proof
{ The proof is provided in Appendix \ref{ProofLemma4}. 
}
\endproof

Thus, we can without loss of generality, assume that ${\bf L}(\mathscr{U}_i)$ are left-orthogonal for all $i$. Then, the projection of a data point ${\bf y}\in \mathbb{R}^{R_n }$ on the subspace ${\bf S_{\text{TT}}}$  is given by 
${\bf L}(\mathscr{U}_1\times_3^{1} \mathscr{U}_2 \times \cdots\times \mathscr{U}_n )^\top {\bf y}$.
\section{Tensor Train Neighborhood Preserving Embedding (TTNPE)} \label{NPE}
Given a set of tensor data $\mathscr{X}_{i}\in \mathbb{R}^{I_1 \times \cdots \times I_n}$,  $i=1,\cdots, N$, 
we wish to project the data  $\mathscr{X}_{i}$ to vector ${\bf t}_i \in \mathbb{R}^{R_n}$, satisfying 
{${\bf t}_i  ={\bf L}(\mathscr{U}_1\times_3^{1} \mathscr{U}_2 \times \cdots \times \mathscr{U}_n )^T {\bf V}(\mathscr{X}_{i}) $}
and preserving neighborhood among the projected data. 
We first construct a {neighborhood} graph  to capture the neighborhood information in the given data and  generate the affinity matrix ${\bf F}$ as
{
\begin{equation}
{\bf F}_{ij} =\left\{
\begin{array}{ll}
      \exp(-\|\mathscr{X}_i -\mathscr{X}_j\|_F^2/ \epsilon), & \text{if } \mathscr{X}_{j: j\neq i} \in O(K, \mathscr{X}_i)\\
      0, & \text{otherwise},
\end{array} 
\right.
\end{equation}
}
where $O(K, \mathscr{X}_i)$ denotes the subset of data excluding $\mathscr{X}_i$ that are within the $K$-nearest neighbors of $\mathscr{X}_i$, and $\epsilon$ is the scaling factor. By definition, ${\bf F}_{ii}=0$. 
We also note that this is an unsupervised tensor embedding method since the label information is not used in the embedding procedure.
Without loss of generality, we set ${\bf S} = {\bf F} +{\bf F}^\top$ and {${\bf S}$ is further normalized by dividing entries of each row by the row sum such that each row sums to one}.

The goal is to find the decomposition $\mathscr{U}_1, \cdots, \mathscr{U}_n$ that minimizes the average distance between all the points and their weighted combination of remaining points, weighted by the symmetrized affinity matrix in the projection, i.e.
{
\begin{equation}\label{eq: TTNPE1}
\begin{split}
\min_{{\substack{ \mathscr{U}_k:  \forall {k=1,\cdots, n} \\  {\bf L}(\mathscr{U}_k)  \text{ is Unitary}}}}
&\sum_i \|{\bf L}^\top(\mathscr{U}_1\times  \cdots \times \mathscr{U}_n  ){\bf V}(\mathscr{X}_i )\\
&- \sum_j {\bf S}_{ij} {\bf L}^\top(\mathscr{U}_1\times  \cdots \times \mathscr{U}_n ){\bf V}(\mathscr{X}_j)\|_2^2.
\end{split}
\end{equation}
}
Let ${\bf D}\in \mathbb{R}^{I_1  \cdots  I_n\times N}$ be the matrix that concatenates the $N$ vectorized tensor data such that the $i^{\text{th}}$ column of ${\bf D}$ is  ${\bf V}({\mathscr{X}}_i)$, and let 
${\bf E}= {\bf L}( \mathscr{U}_1\times  \cdots \times \mathscr{U}_n )$. 
Then,  \eqref{eq: TTNPE1} is equivalent to 
\begin{equation}\label{eq: TTNPE2}
\begin{split}
\min_{\substack{ \mathscr{U}_k: \forall {k=1,\cdots, n} \\ {\bf L}(\mathscr{U}_k)  \text{ is Unitary}}} 
\|{\bf E}^\top ({\bf D} -  {\bf D}{\bf S}^\top)\|_F^2.
\end{split}
\end{equation}
Since ${\bf D} -  {\bf D}{\bf S}^\top \in \mathbb{R}^{(I_1\cdots I_n) \times N}$ is determined,  we set ${\bf Y} = {\bf D} -  {\bf D}{\bf S}^\top$. Thus the Frobenius norm in \eqref{eq: TTNPE2} can be further expressed in the form of matrix trace to reduce the problem to
\begin{equation}\label{eq: TTNPE3}
\min_{\substack{ \mathscr{U}_k: \forall {k=1,\cdots, n} \\ {\bf L}(\mathscr{U}_k)  \text{ is Unitary}}} 	\text{tr}  ( {\bf Y}^\top 
{\bf E}{\bf E}^\top
{\bf Y}) .
\end{equation}
Based on the cyclic permutation property of the trace operator, \eqref{eq: TTNPE3} is equivalent to 
\begin{equation}\label{eq: TTNPE3_5}
\min_{\substack{ \mathscr{U}_k: \forall {k=1,\cdots, n} \\ {\bf L}(\mathscr{U}_k)  \text{ is Unitary}}}
\text{tr}  ( 
{\bf E}^\top {\bf Y}{\bf Y}^\top {\bf E}
). 
\end{equation}
Let ${\bf Z} = {\bf Y} {\bf Y}^\top \in \mathbb{R}^{(I_1\cdots I_n) \times (I_1\cdots I_n)}$ be the constant matrix. Then, the problem \eqref{eq: TTNPE3_5} becomes 
\begin{equation}\label{eq: TTNPE4}
\min_{\substack{ \mathscr{U}_k: \forall {k=1,\cdots, n} \\ {\bf L}(\mathscr{U}_k)  \text{ is Unitary}}} 	 \text{tr}  (  {\bf E}^\top{\bf Z} {\bf E}) .
\end{equation}
We will use the alternating minimization method \cite{beck2015convergence} to solve \eqref{eq: TTNPE4} such that each $\mathscr{U}_k$ is updated by solving
\begin{equation}\label{eq: target0}
\min_{\substack{\mathscr{U}_k: {\bf L}(\mathscr{U}_k)  \text{ is unitary}}} 	 \text{tr}  (  {\bf E}^\top{\bf Z} {\bf E}).
\end{equation}
In order to solve \eqref{eq: target0}, we use an iterative algorithm. Each $\mathscr{U}_{k:k=1,\cdots, n}$ is initialized by tensor train decomposition \cite{oseledets2011tensor} with a thresholding parameter $\tau$, 
{ which zeros out the singular values which are smaller than $\tau$ times the maximum singular value, }
such that tensor train ranks $(R_1,\cdots, R_n)$ are determined. 
{ The larger the thresholding parameter $\tau$, the smaller the tensor train ranks. Typically, $\tau$ could be chosen via cross validation such that the classification error in the validation set is minimized.}

\subsection{Tensor Train Neighbor Preserving Embedding using Tensor Network (TTNPE-TN)}
Let $\mathscr{Z} \in \mathbb{R}^{I_1 \times \cdots \times I_n \times I_1 \times \cdots \times I_n}$ be the reshaped tensor of ${\bf Z}$, and 
{
\begin{equation}
\begin{split}
\mathscr{T}_1 = \mathscr{U}_1 \times \cdots \times \mathscr{U}_{k-1} \in \mathbb{R}^{I_1\times \cdots \times I_{k-1} \times R_{k-1}}, \\
\mathscr{T}_n = \mathscr{U}_{k+1} \times \cdots \times \mathscr{U}_{n} \in \mathbb{R}^{R_{k} \times I_{k+1}\times \cdots \times I_{n} \times R_{n}}.
\end{split}
\end{equation}
}

For {\bf Updating $\mathscr{U}_{k:k=1,\cdots,n-1}$}, based on Lemma \ref{equivalence}, we note that \eqref{eq: target0} can be written as
\begin{eqnarray}
&&\min_{\substack{\mathscr{U}_k \\ {\bf L}(\mathscr{U}_k)  \text{ is unitary}}} 	 
\mathscr{U}_k
{\times_{1,2,3}^{1,2,3}}
  {\text{tr}_{4}^{8}} 
 \left( 
 \mathscr{Z} 		
 \times_{n+k+1, \cdots, 2n}^{2, \cdots, n-k+1} \mathscr{T}_n 		\right.\nonumber\\&&\left.\times_{n+1, \cdots, n+k-1}^{1, \cdots, k-1} \mathscr{T}_1
\times_{k+1, \cdots, n}^{2, \cdots, n-k+1} \mathscr{T}_n 			\times_{1, \cdots, k-1}^{1, \cdots, k-1} \mathscr{T}_1 
\right)\nonumber\\&&
{ \times_{1,2,3}^{1,2,3}} 
\mathscr{U}_k.\label{eq: target05}
\end{eqnarray}
Let $\mathscr{A} \in \mathbb{R}^{R_{k-1} \times I_k \times R_k \times R_{k-1} \times I_k \times R_k}$ be the $6$-order tensor, given as 
$
 {\text{tr}_{4}^{8}} 
 \left( 
 \mathscr{Z} 		
 \times_{n+k+1, \cdots, 2n}^{2, \cdots, n-k+1} \mathscr{T}_n \right.$		$\times_{n+1, \cdots, n+k-1}^{1, \cdots, k-1} \mathscr{T}_1
\times_{k+1, \cdots, n}^{2, \cdots, n-k+1} \mathscr{T}_n $ $\left.			\times_{1, \cdots, k-1}^{1, \cdots, k-1} \mathscr{T}_1 
\right)
$, 
{ where the details to compute $\mathscr{A}$ via tensor merging product is given in Appendix \ref{TNM}.}
Thus \eqref{eq: target05} becomes
{
\begin{equation}\label{eq: target1}
\min_{\substack{\mathscr{U}_k: {\bf L}(\mathscr{U}_k)  \text{ is unitary}}} 	\mathscr{U}_k \times_{1,2,3}^{1,2,3} \mathscr{A} \times_{1,2,3}^{1,2,3} \mathscr{U}_k.
\end{equation}}

Based on Lemma \ref{equivalence}, the tensor merging product \eqref{eq: target1} can be transformed into matrix product. Thus, \eqref{eq: target1} becomes
\begin{equation}\label{eq: target2}
\min_{\substack{\mathscr{U}_k: {\bf L}(\mathscr{U}_k)  \text{ is unitary}}} 	{\bf V}(\mathscr{U}_k)^\top {\bf A} {\bf V} (\mathscr{U}_k),
\end{equation}
where ${\bf A} \in \mathbb{R}^{{ (R_{k-1}I_kR_k) \times  (R_{k-1}I_kR_k)}}$ is the reshaped form of $\mathscr{A}$. 
A differentiable function under unitary constraint can be solved by the algorithm proposed in \cite{wen2013feasible}. In problem \eqref{eq: target2}, the gradient of objective function to ${\bf V}(\mathscr{U}_k)$ is $2 {\bf A} {\bf V} (\mathscr{U}_k)$.

{\bf Updating $\mathscr{U}_n$} is different from solving $\mathscr{U}_{k:k=1,\cdots, n-1}$ since the trace operation
merges the tensor $\mathscr{U}_n$ with itself, thus \eqref{eq: target1} does not apply for solving $\mathscr{U}_n$.
Instead, updating $\mathscr{U}_n$ in \eqref{eq: target0} is equivalent to solving
\begin{eqnarray}
&&\min_{\substack{\mathscr{U}_n: {\bf L}(\mathscr{U}_n)  \text{ is unitary}}} 
\text{tr}_{1}^{2}
\left(
\mathscr{U}_n 
{ \times_{1,2}^{1,2}}
(\mathscr{Z} \times_{n+1, \cdots, 2n-1}^{1, \cdots, n-1} \mathscr{T}_1 \right.\nonumber\\&&\left.\times_{1, \cdots, n-1}^{1, \cdots, n-1}  \mathscr{T}_1 ) 
{ \times_{1,2}^{1,2}}
\mathscr{U}_n
\right).\label{eq:tn}
\end{eqnarray}

Let $\mathscr{B} \in \mathbb{R}^{R_{n-1} \times I_n \times R_{n-1} \times I_n}$ be the $4$-th order tensor formed by 
$
(\mathscr{Z} \times_{n+1, \cdots, 2n-1}^{1, \cdots, n-1} \mathscr{T}_1 \times_{1, \cdots, n-1}^{1, \cdots, n-1}  \mathscr{T}_1 ) ,
$
{ where the details to compute $\mathscr{B}$ via tensor merging product is given in Appendix \ref{TNM2}.}
Thus updating $\mathscr{U}_n$ is equivalent to solving
\begin{equation}
\min_{\substack{\mathscr{U}_n: {\bf L}(\mathscr{U}_n)  \text{ is unitary}}} 
\text{tr}_{1}^{2}
\left(
\mathscr{U}_n 
{ \times_{1,2}^{1,2}}
\mathscr{B}
{ \times_{2,3}^{1,2}}
\mathscr{U}_n
\right),
\end{equation}
which by Lemma \ref{equivalence}, can be transformed into the matrix form
\begin{equation}\label{eq: target3}
\min_{\substack{\mathscr{U}_n \in \mathbb{R}^{R_{n-1} \times I_n \times R_n}\\{\bf L}(\mathscr{U}_n)  \text{ is unitary}}} 	
\text{trace}( {\bf L} \left( \mathscr{U}_n)^\top {\bf B} {\bf L} (\mathscr{U}_n) \right),
\end{equation}
where ${\bf B} \in \mathbb{R}^{(R_{n-1}I_n) \times (R_{n-1}I_n)}$ is reshaped from $\mathscr{B}$. 
The gradient of the objective function to ${\bf L}(\mathscr{U}_n)$ is $2{\bf B} {\bf L} (\mathscr{U}_n) $.

We now analyze the {\bf computation and memory complexity} of TTNPE-TN algorithm, { where the memory complexity indicates the memory required to store all the intermediate variables}.
For $\mathscr{U}_{k:k=1,\cdots n-1}$, the generation of ${\bf A}$ requires merging the tensor networks, which has a computation complexity of 
$O\left( (I_1\cdots I_n)^2R_{k-1} + (I_1\cdots I_n)^2 (\frac{R_{k-1}}{I_1\cdots I_{k-1}})^2 R_k R_n \right)$, and  solving \eqref{eq: target2} takes $O \left( R_{k-1}I_kR_k^2 \right)$ time. Thus, the computation of ${\bf A}$ dominates the complexity. 
The memory requirement for generating ${\bf A}$ is $O\left( (R_{k-1}I_kR_kR_n)^2 \right)$, which is large when the tensor train ranks are high.
Similarly, the generation of ${\bf B}$ to solve $\mathscr{U}_n$ takes $O\left( (I_1\cdots I_n)^2 R_{n-1} \right)$ time and solving \eqref{eq: target3} takes $O(R_{n-1}I_nR_n^2)$, and the memory for generating ${\bf B}$ is $O\left( (I_1 \cdots I_n)^2 \right)$, indicating solving for $\mathscr{U}_n$ is less expensive than that for solving for $\mathscr{U}_k$ in terms of both memory and computation complexity. 

Although TTNPE-TN algorithm gives an exact solution for updating $\mathscr{U}_i$ in each alternating minimization step, the  memory and computation cost prohibits its application when the tensor train ranks are large. 
In order to address this,  we propose a Tensor Train  Neighbor Preserving Embedding using Approximate Tensor Network (TTNPE-ATN) algorithm in the next section, to approximate \eqref{eq: target0}, aiming to reduce computation and memory cost.

\subsection{Tensor Train Neighbor Preserving Embedding using Approximated Tensor Network (TTNPE-ATN)}
Our main intuition is as follows. Without the TT decomposition constraint, the solution to minimize the quadratic form  $ \text{tr}  (  {\bf E}^\top{\bf Z} {\bf E}) $ where ${\bf E}$ is unitary is given by ${\bf E}$ being the matrix formed by eigenvectors corresponding to the lowest eigenvalues of ${\bf Z}$ and the value of the objective is the sum of the lowest eigenvalues of ${\bf Z}$ \cite{moslehian2012ky}. Let the matrix corresponding to the eigenvectors corresponding to $r_n$ {smallest} eigenvalues of ${\bf Z}$ be ${\bf V}_{r_n}$. 
With the additional constraint that ${\bf E}$ has TT decomposition, the above choice of ${\bf E}$ may not be optimal. 
Thus, we relax the original problem to minimize the distance between ${\bf E}$ and ${\bf V}_{r_n}$. Thus, the relaxed problem of \eqref{eq: target0} is 
{
\begin{equation}\label{eq: TTNPE5}
\min_{\substack{ \mathscr{U}_k  \\ {\bf L}(\mathscr{U}_k)  \text{ is unitary}}} 
\| {\bf L}(\mathscr{U}_1 \times \cdots \times \mathscr{U}_n ) -{\bf V}_{r_n}\|_F^2,
\end{equation}
where 
${\bf L}(\mathscr{U}_1 \times \cdots \times \mathscr{U}_n ), {\bf V}_{r_n} \in \mathbb{R}^{(I_1 \cdots I_n) \times r_n}$. }

Let ${\bf T}_k$ be a reshaping operator that change the dimension of a matrix from $\mathbb{R}^{(I_1\cdots I_n) \times r_n}$ to $\mathbb{R}^{(I_1 \cdots I_k) \times (I_{k+1}\cdots I_n r_n)}$, thus \eqref{eq: TTNPE5} is equivalent to
\begin{equation} \label{eq: TTNPE9}
\min_{\substack{ \mathscr{U}_k : {\bf L}(\mathscr{U}_k)  \text{ is unitary}}} 
\| {\bf T}_k({\bf L}(\mathscr{T}_1 \times_{k}^{1} \mathscr{U}_k \times_{3}^{1} \mathscr{T}_n )) - {\bf T}_k({\bf V}_{r_n})\|_F^2,
\end{equation}
which is equivalent to
\begin{equation}\label{eq: TTNPE9_1}
\min_{\substack{ \mathscr{U}_k: {\bf L}(\mathscr{U}_k)  \text{ is unitary}}} 
\| \left( {\bf I}_{I_k} \otimes {\bf L}( \mathscr{T}_1) \right) 
{\bf L}(\mathscr{U}_k) 
{\bf R}(\mathscr{T}_n)-{\bf T}_k({\bf V}_{r_n})\|_F^2,
\end{equation}
{ which has the same format as minimizing $\|{\bf PXQ} - {\bf C}\|_F^2$ under unitary constraint. Since the gradient  is ${\bf P}^\top ({\bf PXQ-C}){\bf Q}^\top$, \eqref{eq: TTNPE9_1} can be solved by the algorithm proposed in \cite{wen2013feasible}.}

After the relaxation,
the computation complexity is $O\left(  R_{k-1}I_kI_1\cdots I_n R_n\right)$ for calculating the gradient,  $O\left( (I_1\cdots I_n)^2 \right)$ for generating ${\bf V}_{r_n}$, and  $O(R_{k-1}I_kR_k^2)$ for solving \eqref{eq: TTNPE9_1}.
Thus the eigenvalue decomposition for generating ${\bf V}_{r_n}$ dominates the computational complexity.
The memory for computing { ${\bf P}$ and ${\bf Q}$} is $\max(I_1\cdots I_{k-1}R_{k-1}, R_kI_{k+1}\cdots I_n R_n)$. 
Thus both memory and computation cost of TTNPE-ATN are much less than those in TTNPE-TN algorithm. 
Therefore in the simulation section, we will only consider the TTNPE-ATN algorithm. We validated for a small experiment that the embedding performance for the two are similar, where the validation results are omitted in this paper. The two algorithms (TTNPE-TN  and TTNPE-ATN) are described in Algorithm \ref{TTNPE-Algo}. 

\begin{algorithm}
   \caption{TTNPE-TN and TTNPE-ATN Algorithms}
   \begin{algorithmic}[1]
   \INPUT  { A set of $N$ tensors $\mathscr{X}_{i=1,\cdots, N} \in \mathbb{R}^{I_1\times I_2\times \cdots \times I_n }$, denoted as $\mathscr{X}$},  threshold parameter $\tau$, kernel scaling parameter $\epsilon$, number of neighbors $K$, { thresholding parameter $\tau$}, and max iterations $maxIter$
    \OUTPUT  Tensor train subspace factors $\mathscr{U}_1, \mathscr{U}_2, \cdots, \mathscr{U}_n$
    \STATE Compute affinity matrix ${\bf F}$ by
    \[   {\bf F}_{ij} =\left\{
\begin{array}{ll}
      \exp(-\|\mathscr{X}_i -\mathscr{X}_j\|_F^2/ \epsilon), & \text{if } \mathscr{X}_{j: j\neq i} \in O(K, \mathscr{X}_i), \\
      0, & \text{otherwise},
\end{array} 
\right. 
\]
\[
{\bf S} = {\bf F} + {\bf F}^\top \text{ and normalize } {\bf S} \text{ such that each row sums to 1. }
\]

	\STATE Form ${\bf D}\in \mathbb{R}^{I_1  \cdots  I_n\times N}$ as a reshape of the input data, compute ${\bf Y} = {\bf D} -  {\bf D}{\bf S}^\top$, and compute ${\bf Z} = {\bf Y} {\bf Y}^\top$.
	\STATE {Apply tensor train decomposition \cite{oseledets2011tensor} on $\mathscr{X}$ to initialize $\mathscr{U}_{i=1,\cdots, n} $ with thresholding parameter $\tau$, and the tensor train ranks are determined based on selection of $\tau$. }
	\STATE  Solve ${\bf V}_{R_n}$ by applying eigenvalue decomposition on ${\bf Z}$.
	
	\STATE Set $iter =1$ 
	
    	\WHILE { $iter \leq maxIter$ or convergence of $\mathscr{U}_1, \cdots, \mathscr{U}_n$}
		 \FOR{ $i = 1$ to $n$}
		 	\STATE {\bf \text{(TTNPE-TN) }}  Update $\mathscr{U}_{i}$ in \eqref{eq: target2} for  $i< n$ and in  \eqref{eq: target3} for $i=n$, using the algorithm proposed in \cite{wen2013feasible}.
			\STATE {\bf \text{(TTNPE-ATN) }} Update $\mathscr{U}_i$ in \eqref{eq: TTNPE9_1} by algorithm proposed in \cite{wen2013feasible}.
		 \ENDFOR
		\STATE $iter = iter +1 $
		 
	\ENDWHILE

\end{algorithmic}
\label{TTNPE-Algo}
\end{algorithm}

\subsection{Classification Using TTNPE-TN and TTNPE-ATN}
The classification is conducted by first solving a set of {tensor train factors}
$\mathscr{U}_1 ,\cdots,\mathscr{U}_n$.
Then,  
 the training data and testing data is projected  onto the tensor train subspace bases as follows:
 {
\begin{equation}
{\bf t}_i = {\bf L}(\mathscr{U}_1 \times \cdots \times \mathscr{U}_n )^\top {\bf V}(\mathscr{X}_i) \in \mathbb{R}^{R_n}.
\end{equation}
}
Any data point in the testing set is labeled by applying k-nearest neighbors(KNN)\cite{altman1992introduction} classification with $K$ neighbors in the embedded space $\mathbb{R}^{R_n}$. 
\vspace{-.1in}
\subsection{Storage and Computation Complexity}
In this section, we will analyze the amount of storage to store the high dimensional data, complexity for finding the embedding using TTNPE-ATN and the cost of projection onto the TT subspace for classification. 
KNN and TNPE \cite{dai2006tensor} algorithms are considered for comparison.  
For the computational complexity analysis,  let $d$ be the data dimension, $n$ and $r$ be the reshaped tensor order and rank in TTNPE-ATN model, $K$ be the number of neighbors, and $N_{\text{tr}}\text{ }(N_{\text{te}})$ be the total training (testing) data. We assume the dimension along each tensor mode is the same, thus each tensor mode is $d^{\frac{1}{n}}$ in dimension.  
\begin{table*}[t!]
{
  \centering
  \scalebox{.8}{
  \begin{tabular}{cccc}
    \toprule
    & Storage & Subspace Computation & Classification\\
    \midrule
    KNN		&  $dN_{tr}$												& ${\bf 0}$ 								& $O(N_{{te}}  N_{{tr}} d)$\\
    TNPE		&  $r^nN_{tr} + nd^{\frac{1}{n}} r $. 								& $O(n(N_{tr}rd + N_{tr}r^{2n} + r^{3n}))$ 			& $O(N_{{te}}r^2d + N_{{te}}N_{{tr}}r^n)$\\
    TTNPE-ATN 	&  $\boldsymbol{(n-1)(d^{\frac{1}{n}}r^2 -r^2) + (d^{\frac{1}{n}}r - r^2)+ rN_{tr}}$ 	&  $O(nd^{\frac{1}{n}}r^3 + dN_{tr}^2 +d^2N_{tr})$ 	& $\boldsymbol{O(N_{te}r^2d + N_{te}N_{tr}r)}$\\
    \bottomrule
  \end{tabular}
  }
  }
  \caption{\small Storage and Computation Complexity Analysis for  Embedding Methods. The bold entry in each column depicts the lowest order. }
  \centering
   \label{ComplexityTable2}
  \vspace{.2in}
\end{table*}

{\bf Storage of data }
Under KNN model, the storage required for $N_{\text{tr}} $ training data is
$\text{Storage(KNN)} = dN_{\text{tr}}$.
Under TNPE model, the storage for the $N_\text{tr}$ training data needs the space for 
{$n$ linear transformation which is $n(d^{\frac{1}{n}} r)$},
and the space for $N_\text{tr}$ embedded training data of size $N_\text{tr} r^n$, requiring the total storage 
{$\text{Storage(TNPE)} =r^nN_{\text{tr}} + nd^{\frac{1}{n}} r $}.
Under TTNPE-ATN model, we need space 
{ $(n-1)(d^{\frac{1}{n}}r^2 -r^2) + (d^{\frac{1}{n}}r - r^2)$} \cite{holtz2012manifolds}
to store the projection bases $\mathscr{U}_1, \cdots, \mathscr{U}_n$, and $N_\text{tr} r$ to store the embedded training data. Thus the total storage is 
{ $\text{Storage(TTNPE-ATN)} = (n-1)(d^{\frac{1}{n}}r^2 -r^2) + (d^{\frac{1}{n}}r - r^2)+ rN_\text{tr}$.}
We consider a metric of normalized storage, {\em compression ratio}, which is the ratio of storage required under the embedding method and storage for the entire data, calculated by $\rho_{\text{ST}} = \frac{\text{Storage(ST)}}{N_\text{tr}d}$, where $\text{ST}$ can be any of KNN, TNPE, TTNPE-ATN.

{\bf Computation Complexity for estimating the embedding subspace}
{ The computation complexity includes computation for both the addition and multiplication operations. }
Under KNN model, data is directly used for classification and there is no embedding process. 
Under TNPE model, the embedding needs 3 steps, where solving $n$ linear transformations takes $O(N_{\text{tr}}rd)$ for embedding raw data, 
matrix generation for an eigenvalue problem takes $O(N_{\text{tr}}r^{2n})$, 
and eigenvalue decomposition for updating each linear transformation takes $r^{3n}$, 
giving a total computational complexity $O(n(N_{\text{tr}}rd +N_{\text{tr}}r^{2n} + r^{3n}))$.
Under TTNPE-ATN model, the embedding takes 3 steps, 
where the initialization by tensor train decomposition algorithm takes $O(nd^{\frac{1}{n}}r^3)$,
the generation of ${\bf Z}$ takes $O(dN_\text{tr}^2 +d^2N_\text{tr} )$,
and updating $\mathscr{U}_k$, which includes a gradient calculation by merging a tensor network, takes $O(nd^{\frac{1}{n}} r^3))$,
thus giving a total computational complexity $O(nd^{\frac{1}{n}}r^3 + dN_\text{tr}^2 +d^2N_\text{tr})$.

{\bf Classification Complexity}
Under KNN model, classification is conducted by pair-wise computations of the distance between a testing point with all training points, which has a computational complexity of $O(N_{\text{te}}  N_{\text{tr}} d)$.
Under TNPE model, an extra time is required for embedding the testing data, which is  
$O(r^2d N_{\text{te}})$.
However, less time is needed in classification by applying KNN in a reduced dimension, which is $O(N_{\text{tr}}N_{\text{te}}r^n)$. Thus the total complexity is $O(r^2dN_{\text{te}} + N_{\text{tr}}N_{\text{te}}r^n)$.
{
Similarly, under TTNPE-ATN algorithm, embedding takes an extra computation time of  
$O(N_\text{te}r^2d)$,
but a significantly less time used in classification, which is $O(N_\text{te}N_\text{tr}r)$. Thus the total complexity is  $O(N_\text{te}r^2d + N_\text{te}N_\text{tr}r)$.

The comparison of the three algorithms is shown in Table \ref{ComplexityTable2}, where TTNPE-ATN exhibates a great advantage in storage and computation for classification after embedding. 
}

\section{Experiment Results}\label{simu}
\begin{figure*}[t!]
\includegraphics [trim=.1in .2in .1in 1in, keepaspectratio, width=0.32\textwidth] {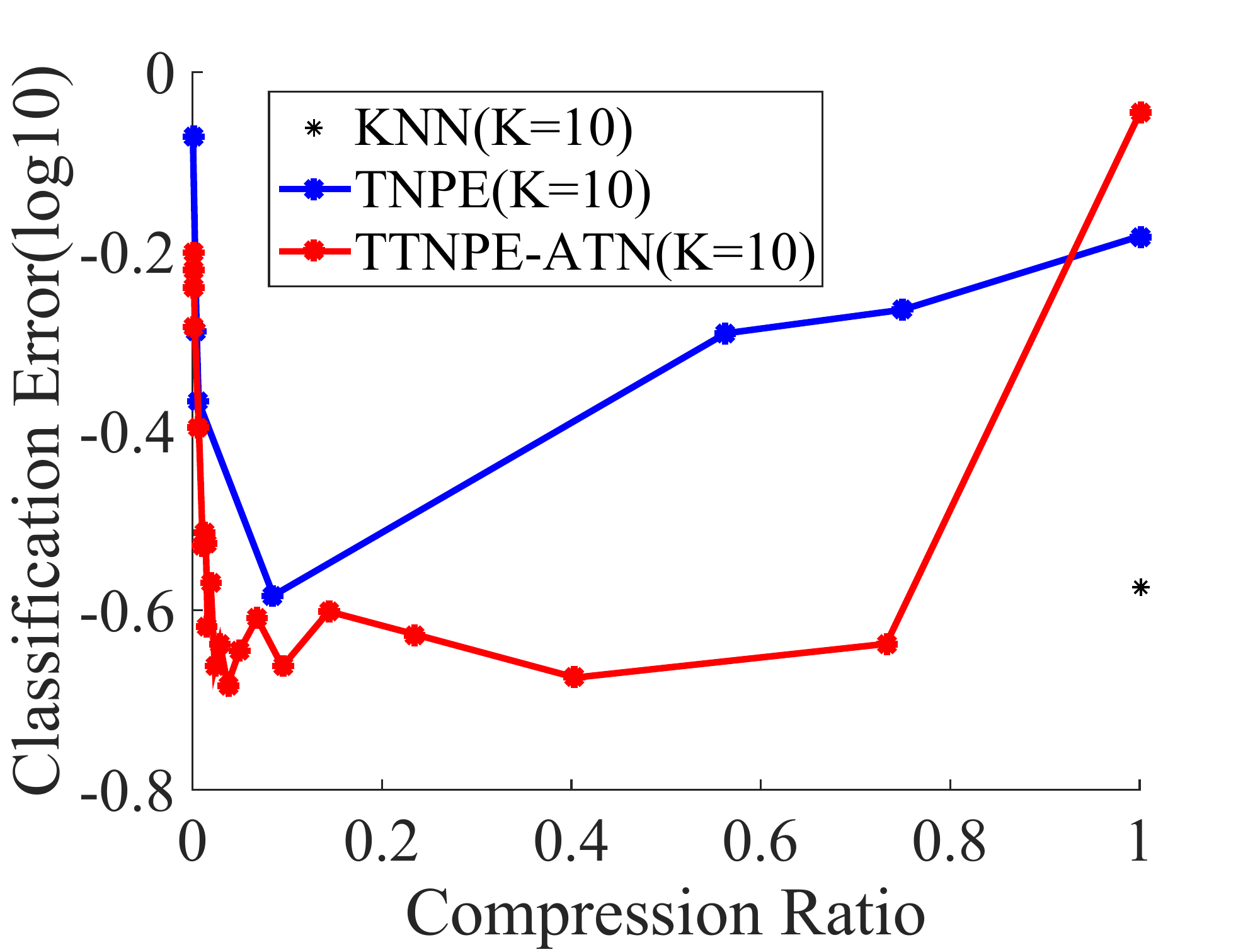}
\includegraphics [trim=.1in .2in .1in 1in, keepaspectratio, width=0.32\textwidth] {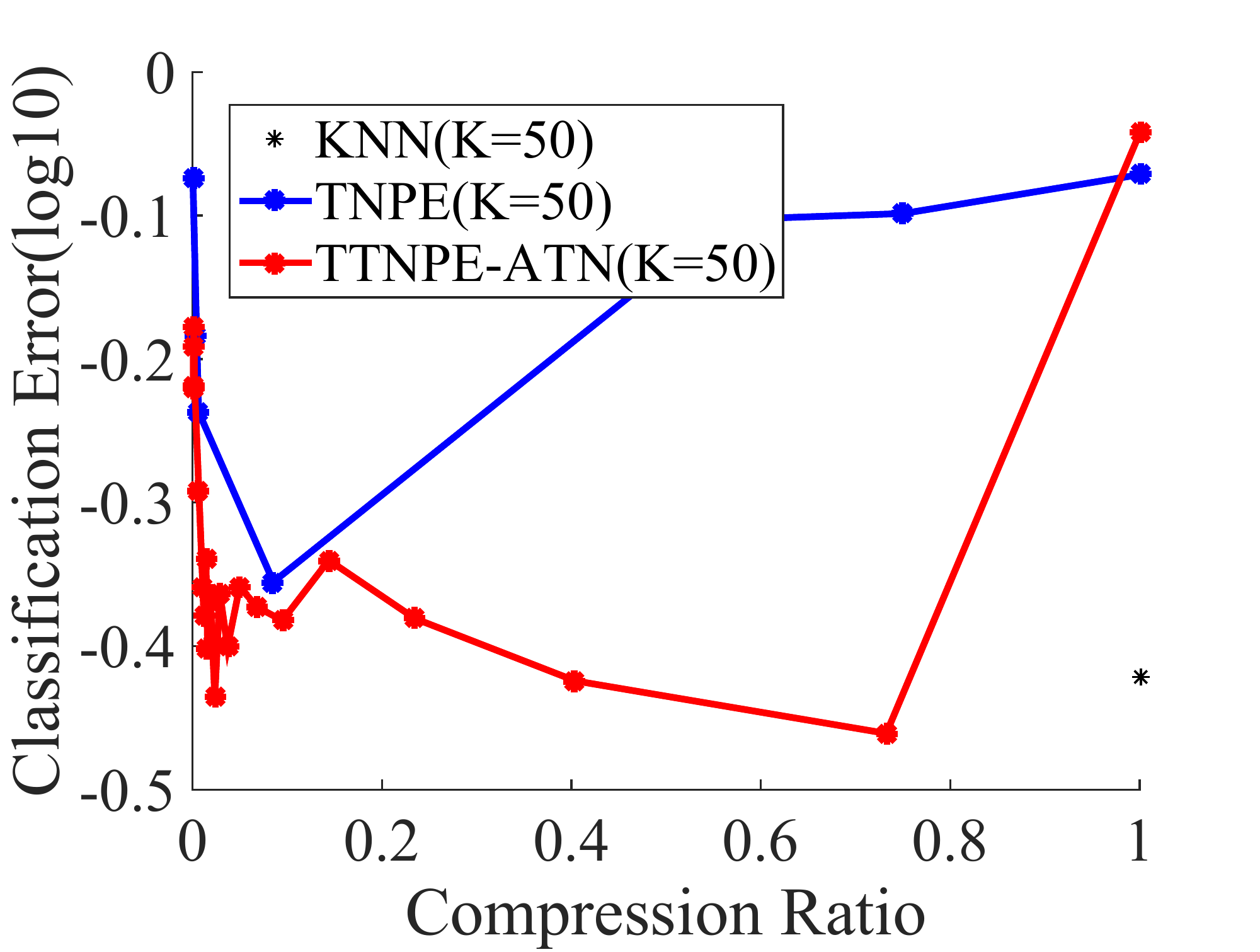}
\includegraphics [trim=.1in .2in .1in 1in, keepaspectratio, width=0.32\textwidth] {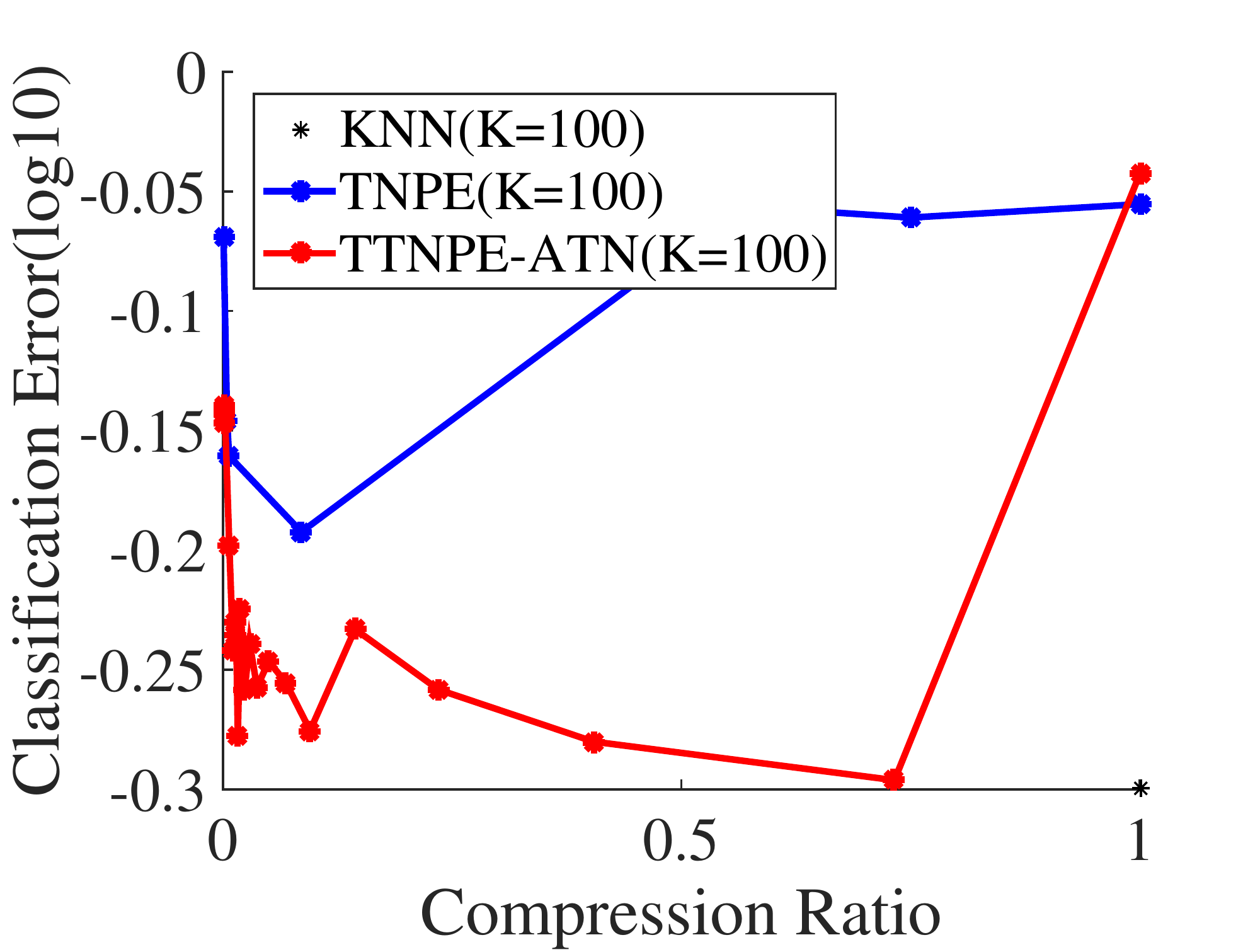}
\centering
\caption{\small {Classification Error in $\log10$ scale for Weizmann Face Database for the three models. } }
\label{TTNPE_Weizmann}
\end{figure*}

In this section, we test our proposed tensor embedding on image datasets, where the  2D images are reshaped into  multi-mode tensors.  Reshaping images to tensors is a common practice to compare tensor algebraic approaches \cite{zhao2016tensor} since it {{captures the low rank property from the data and}} exhibits improved data representation. 
The embedding is evaluated based on KNN classification, where an effective embedding that preserves neighbor information would give classification results close to that of KNN classification at lower compression ratios. 
We compare the proposed  TTNPE-ATN algorithm with Tucker decomposition based  neighbor preserving embedding (TNPE) algorithm as proposed in \cite{dai2006tensor}. We further note that the authors of \cite{dai2006tensor} compared their approach with different approaches based on vectorization of data, including Neighborhood Preserving Embedding (NPE), Locality
Preserving Projection (LPP), { Principal Component Analysis (PCA)}, and Local Discriminant
Embedding (LDE). Since the approach in \cite{dai2006tensor}  was shown to outperform these approaches, we do not consider these vectorized data approaches in our comparison. 
Note that the tensor train rank, which determines the compression ratio, is learnt from the Algorithm \ref{TTNPE-Algo} based upon the selection of $\tau \in (0,1]$.

\subsection{Weizmann Face Database}
Weizmann Face Database \cite{Weizmann} is a dataset that includes 26 human faces with different expressions and lighting conditions. 
66 images from each of the 10 randomly selected people are used for multi-class classification, where 20 images from each person are selected for training and the remaining images are used for testing. 
The experiment is repeated 10 times (for the same 10 people, but random choices of the 20 training images per person) and the averaged classification errors are shown in Fig. \ref{TTNPE_Weizmann}.  
Each image is  down sampled to  $64 \times 44$ for ease of computation and is further reshaped  to a $5$-mode tensor of dimension $4\times 4 \times 4 \times 4 \times 11$ to apply the TNPE and TTNPE-ATN algorithms. 
10, 50, and 100 neighbors are considered to build the graph (from left to right) and the KNN from the same number of neighbors in the embedded space are used for classification. Since KNN does not compress the data, it results in a single point at  a compression ratio of $1$.

We show that TTNPE-ATN  performs better than TNPE when the  compression ratio is lower than $0.9$, indicating TTNPE-ATN better captures the localized features in the dataset thus yielding better embedding  under low compression ratios. 
With the increase of compression ratio, the classification error for TTNPE-ATN algorithm first decreases, which is because the data structure can be better captured with increasing compression ratio (lower compression).  The classification error then increases with compression ratio since the embedding overfits the background noise in the images. Similar trend happens for TNPE algorithm. We note that for a compression ratio of 1, the result for TTNPE-ATN do not match that of KNN since we are learning at-most 200-rank space (due to 20 training images for each of 10 people) while the overall data dimension is $64\times44$, thus giving an approximation at the compression ratio of $1$. 
 Increasing $K$ helps preserve more neighbors for embedding, and the neighbor structure is preserved better. 
Further, the best classification results given by TTNPE-ATN are  even better than the classification results given by KNN algorithm,   
indicating TTNPE-ATN gives better neighborhood preserving embedding as compared to the TNPE algorithm. 

{
{\bf Reshaping} is investigated to verify if the performance of the embedding is subject to the empirically selected reshaping dimension ($4\times 4 \times 4 \times 4 \times 11$). 
The optimal reshaping dimension has been empirically investigated in \cite{wang2017tensor}, where a moderate reshaping gives the best data representation of the multi-dimensional data. Fig. \ref{RS_WM} considers two of the possible reshapings, $4\times 4 \times 4 \times 4 \times 11$ and $8\times 8 \times  4 \times 11$, and illustrates that both TTNPE-ATN and TNPE are not very sensitive to the reshaping method. Further, TTNPE-ATN performs better than TNPE in both the considered reshaping scenarios.
\begin{figure}[ht]
\includegraphics [trim=.1in .2in .1in .1in, keepaspectratio, width=0.45\textwidth] {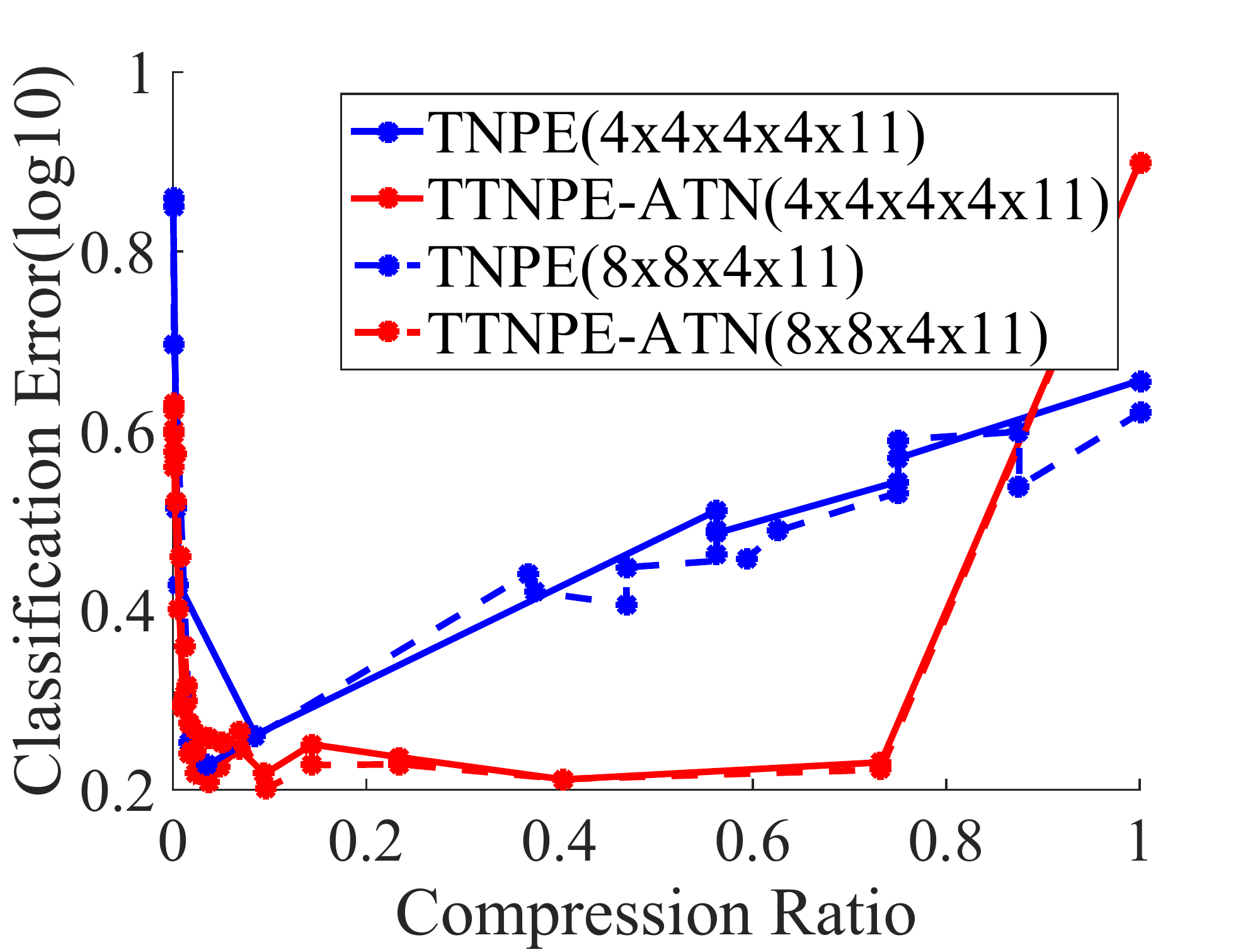}
\centering
\caption{\small Classification Error in $\log10$ scale for Weizmann dataset under reshaping $4\times 4\times 4\times 4\times 11$ and $8\times 8 \times 4\times11$.  }
\label{RS_WM}
\end{figure}

{\bf Noise} perturbation has been investigated for TTNPE-ATN algorithm in Fig. \ref{noise}, where 20dB, 15dB, 10dB, and 5dB  Gaussian noise is added to the data. The performance of TTNPE-ATN algorithm downgrades when the noise increases, while TTNPE-ATN still out-performs than TNPE on clean data when noise is less then 10dB.
\begin{figure}[ht]
\includegraphics [trim=.1in .2in .1in .1in, keepaspectratio, width=0.45\textwidth] {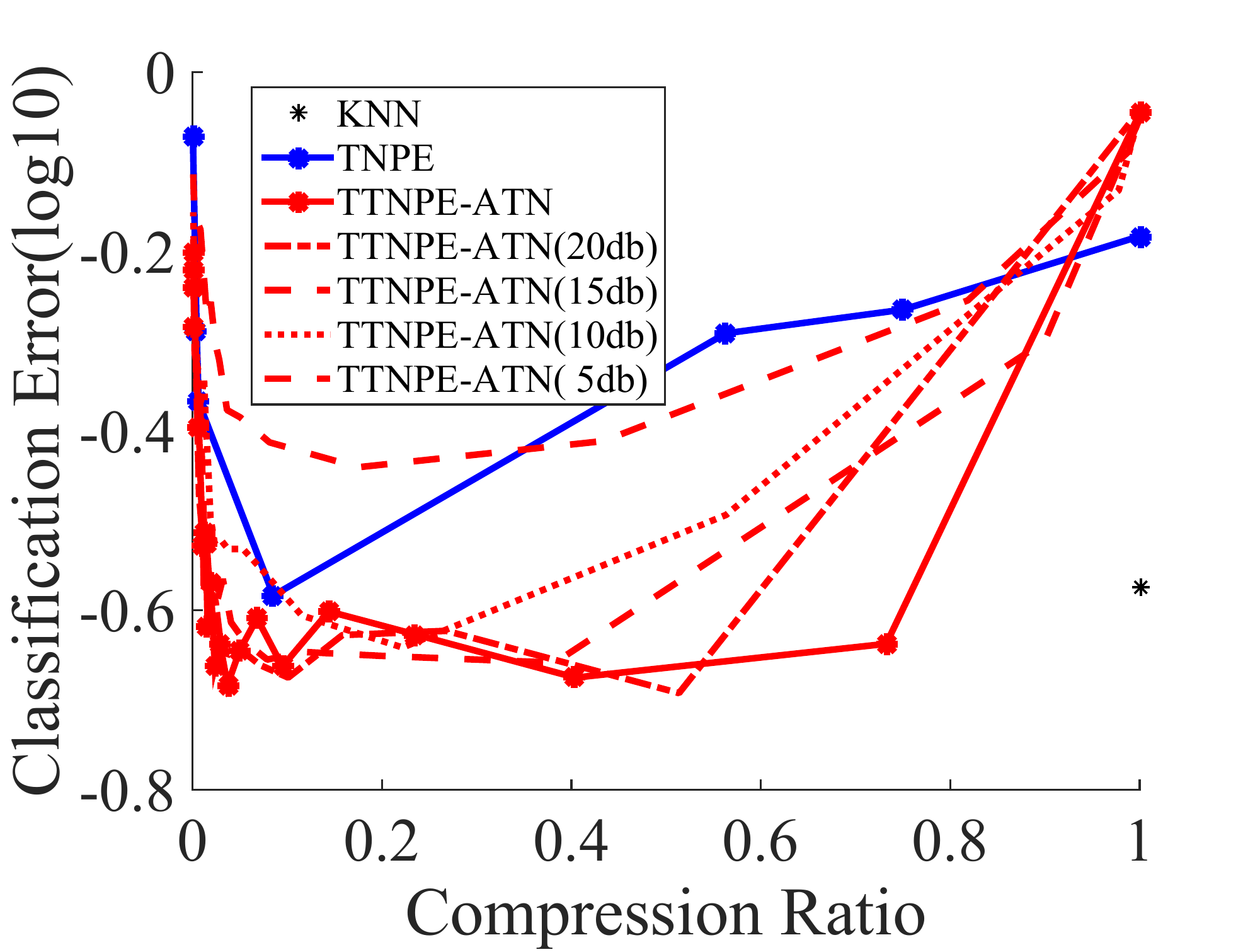}
\centering
\caption{\small Classification Error in $\log10$ scale for Weizmann dataset under noise level 20dB, 15dB, 10dB, and 5dB.  }
\label{noise}
\end{figure}

{\bf Execution time}  for tensor embedding on Weizmann dataset is illustrated in Fig. \ref{time}, where we see that the proposed TTNPE-ATN is faster than TNPE in all of subspace learning, multi-dimensional data embedding, and embedded data classification operations. We also note the time for subspace learning dominates the computation time. Further, the summation of embedding time and classification time is also lower for TTNPE-ATN as compared to KNN. 

\begin{figure*}[t!]
\includegraphics [trim=.1in .2in .1in .1in, keepaspectratio, width=0.32\textwidth] {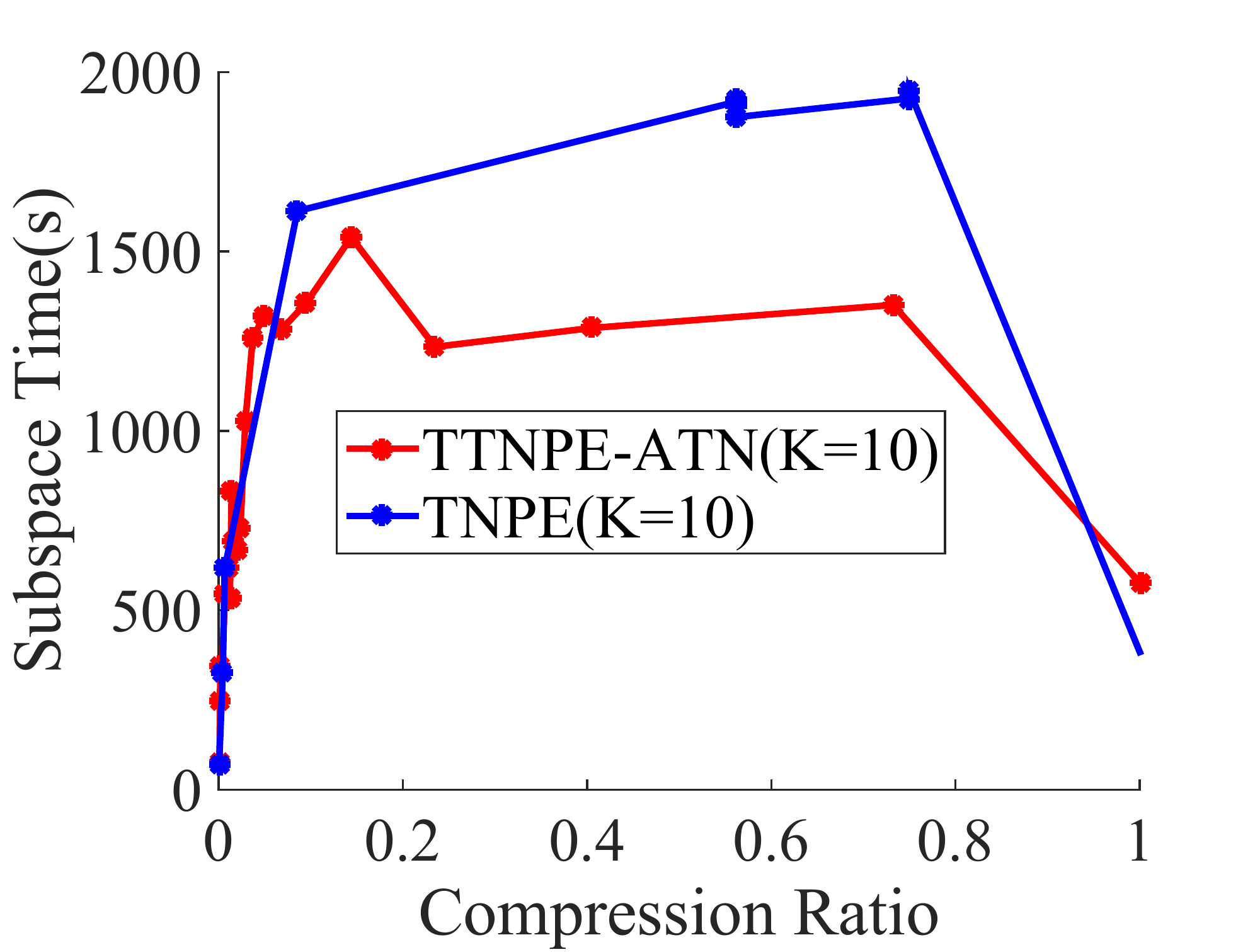}
\includegraphics [trim=.1in .2in .1in .1in, keepaspectratio, width=0.32\textwidth] {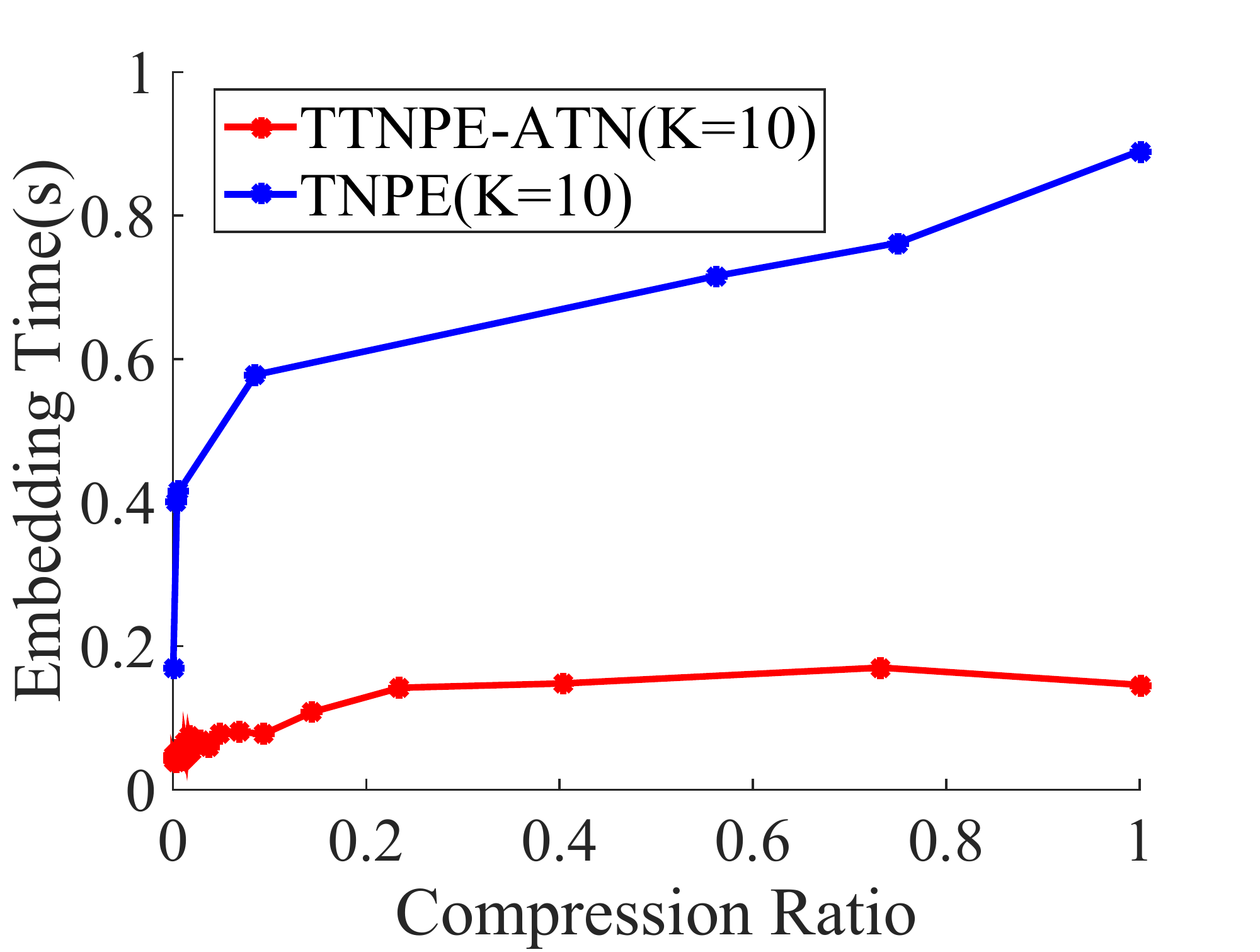}
\includegraphics [trim=.1in .2in .1in .1in, keepaspectratio, width=0.32\textwidth] {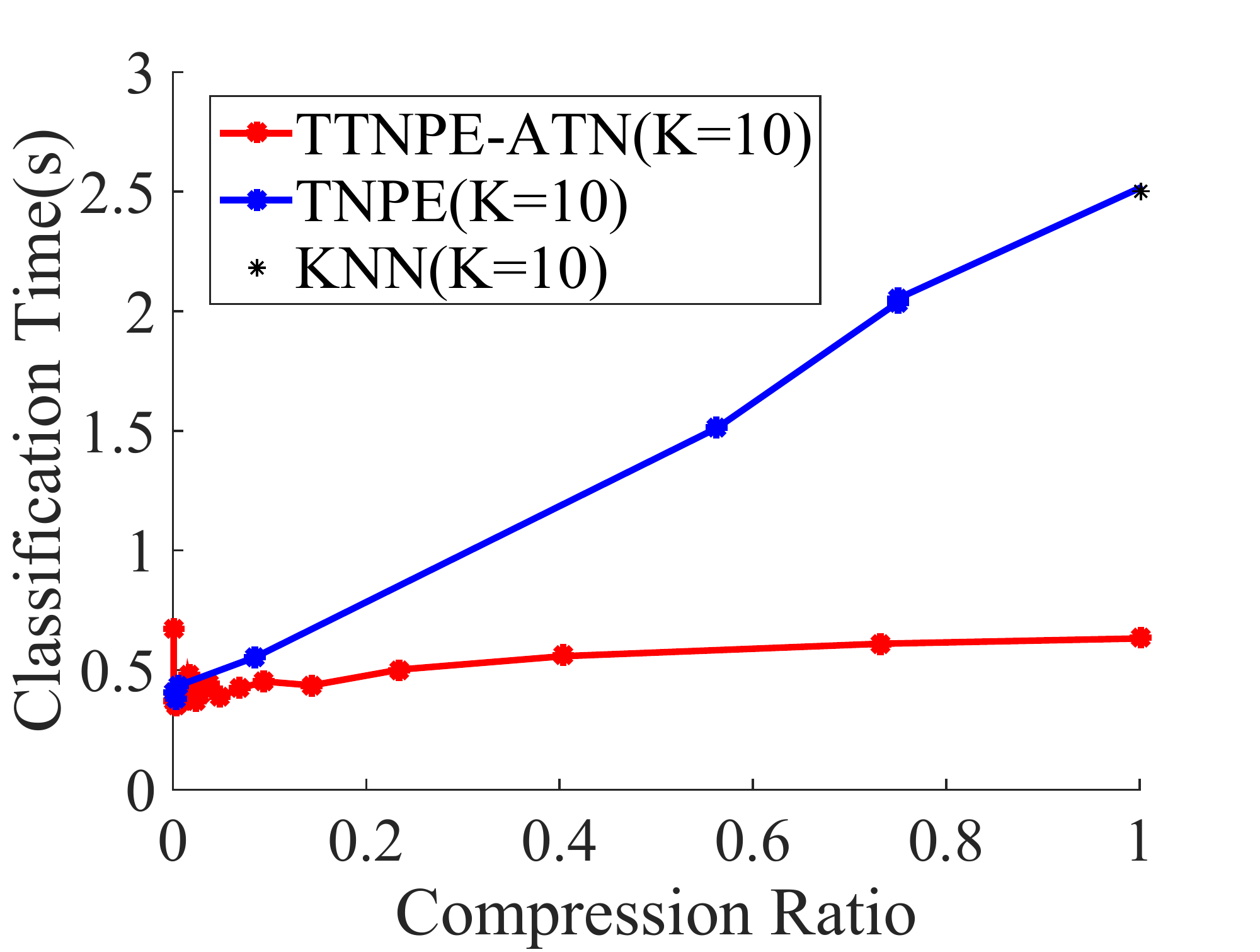}
\centering
\caption{\small CPU time for TNPE, TTNPE-ATN and KNN on Weizmann dataset. From left to right are cpu time for subspace learning, multi-dimensional data embedding, and embedded data classification. The execution time is analyzed using Weizmann dataset when $K$ is chosen to be $10$}
\label{time}
\end{figure*}
}

\subsection{MNIST Dataset}
\begin{figure*}[t!]
\includegraphics [trim=.1in .2in .1in .1in, keepaspectratio, width=0.32\textwidth] {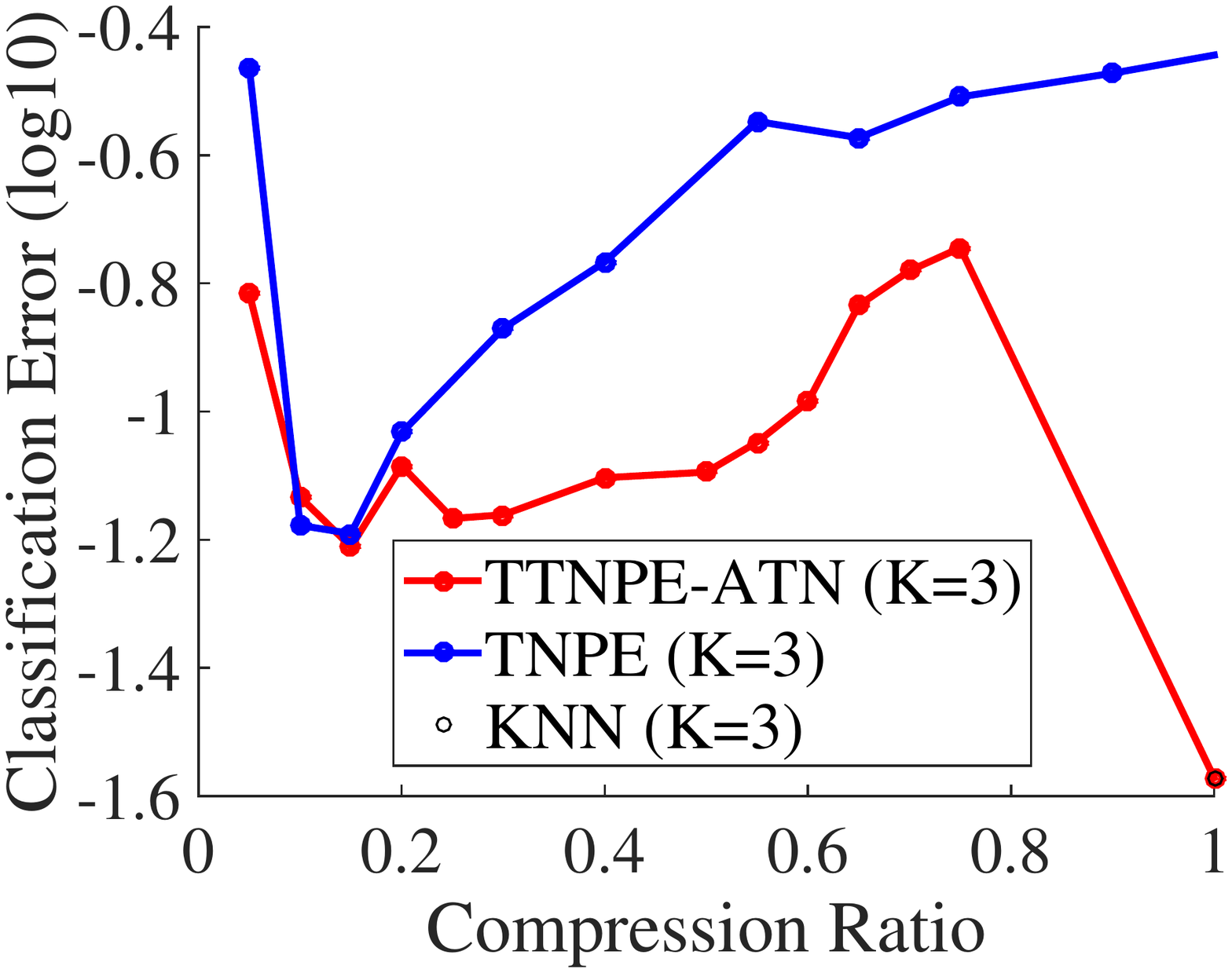}
\includegraphics [trim=.1in .2in .1in .1in, keepaspectratio, width=0.32\textwidth] {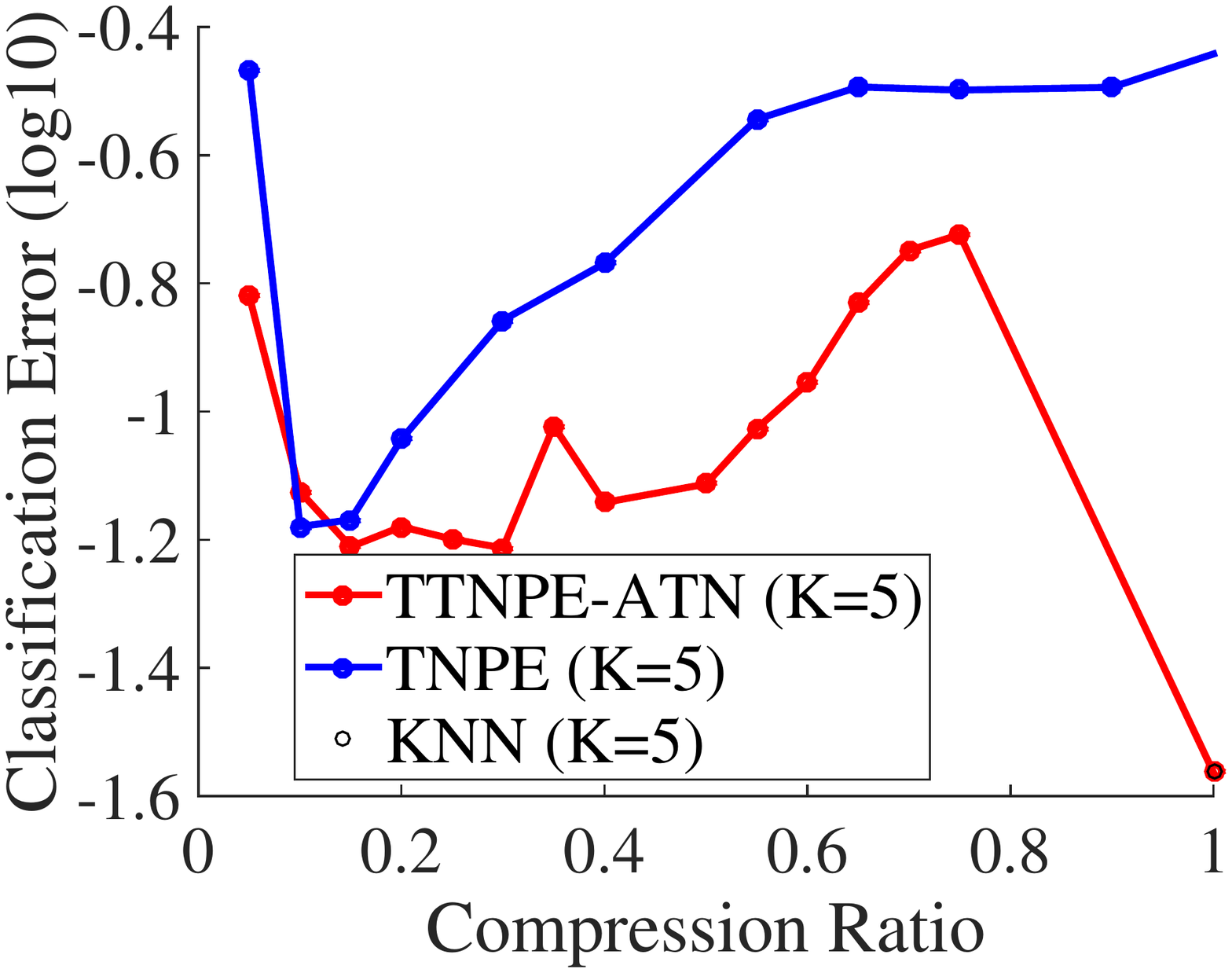}
\includegraphics [trim=.1in .2in .1in .1in, keepaspectratio, width=0.32\textwidth] {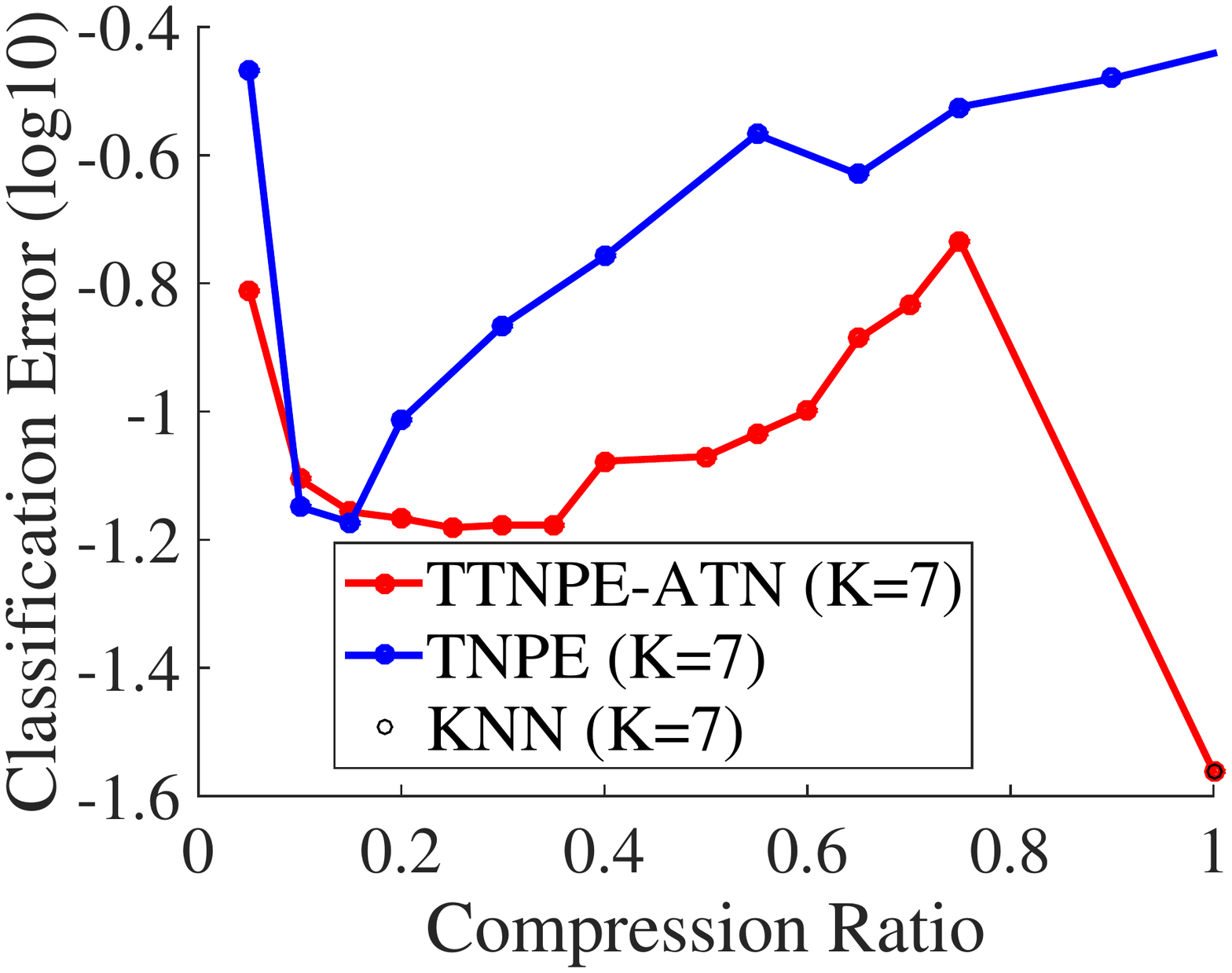}
\centering
\caption{\small Classification Error in $\log10$ scale for MNIST for the three models.  }
\label{TTNPE_MNIST}
\end{figure*}

We use the MNIST dataset \cite{lecun1998gradient}, which consists 60000  handwritten digits of size $28 \times 28$ from $0$ to $9$, to further investigate the embedding performance when the number of training samples is  large.  
Each image is  reshaped to $4\times 7 \times 4 \times 7$ tensor. We perform binary classification for digits $1$ and $2$ by using $600$ training samples from each digit. 
Figure \ref{TTNPE_MNIST} shows the classification performance of the three algorithms (KNN on data directly, TNPE, and TTNPE-ATN) when different values of $K =3,5, 7$ neighbors are used to construct the graph (from left to right). The same value of $K$ is used for classification in the embedded space. 1000 out of sample images from each digit are selected for testing.
  The results in Fig. \ref{TTNPE_MNIST} are averaged over 10 independent experiments (over the choice of 600 training and 1000 test samples). 

We first note that the proposed TTNPE-ATN is the same as the standard KNN for that point when the training sample size is sufficient large (since the number of training samples do not limit the performance). Further, as the compression ratio increases, the classification error of the proposed TTNPE-ATN decreases first, since TTNPE-ATN model can  effectively capture  the embedded data structure. The  classification error then increases since it fits the inherent noise as compared to the low TT-rank approximation of the data. {\em Overall, TTNPE-ATN  algorithm shows comparable embedding performance as TNPE algorithm in the compression ratio region around 0.1, outperforms TNPE for higher compression ratios (lesser compression), and converges to KNN results at compression ratio of 1.}

{We note that TTNPE-ATN shows a different behavior for compression ratios close to 1 in Fig. \ref{TTNPE_MNIST} as compared to Fig. \ref{TTNPE_Weizmann}. This is in part since the number of training samples are lower than the dimension of the data in Fig. \ref{TTNPE_Weizmann} which implies there is an overfitting of noise, while the number of training samples are higher than the data dimension for the results in Fig. \ref{TTNPE_MNIST}.}

{
\subsection{Financial Market Dataset}
\begin{figure*}[t!]
\includegraphics [trim=.1in .2in .1in .1in, keepaspectratio, width=0.32\textwidth] {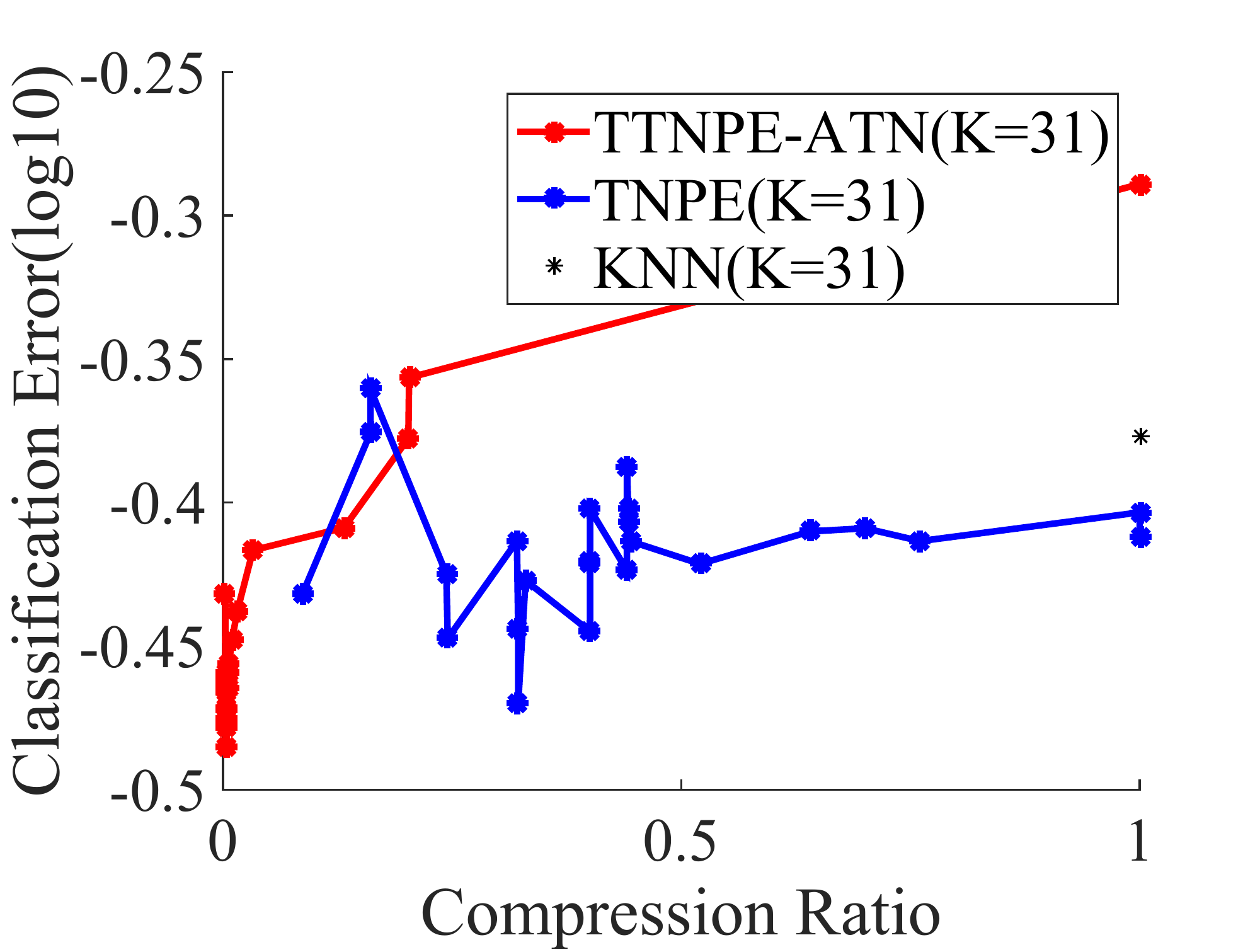}
\includegraphics [trim=.1in .2in .1in .1in, keepaspectratio, width=0.32\textwidth] {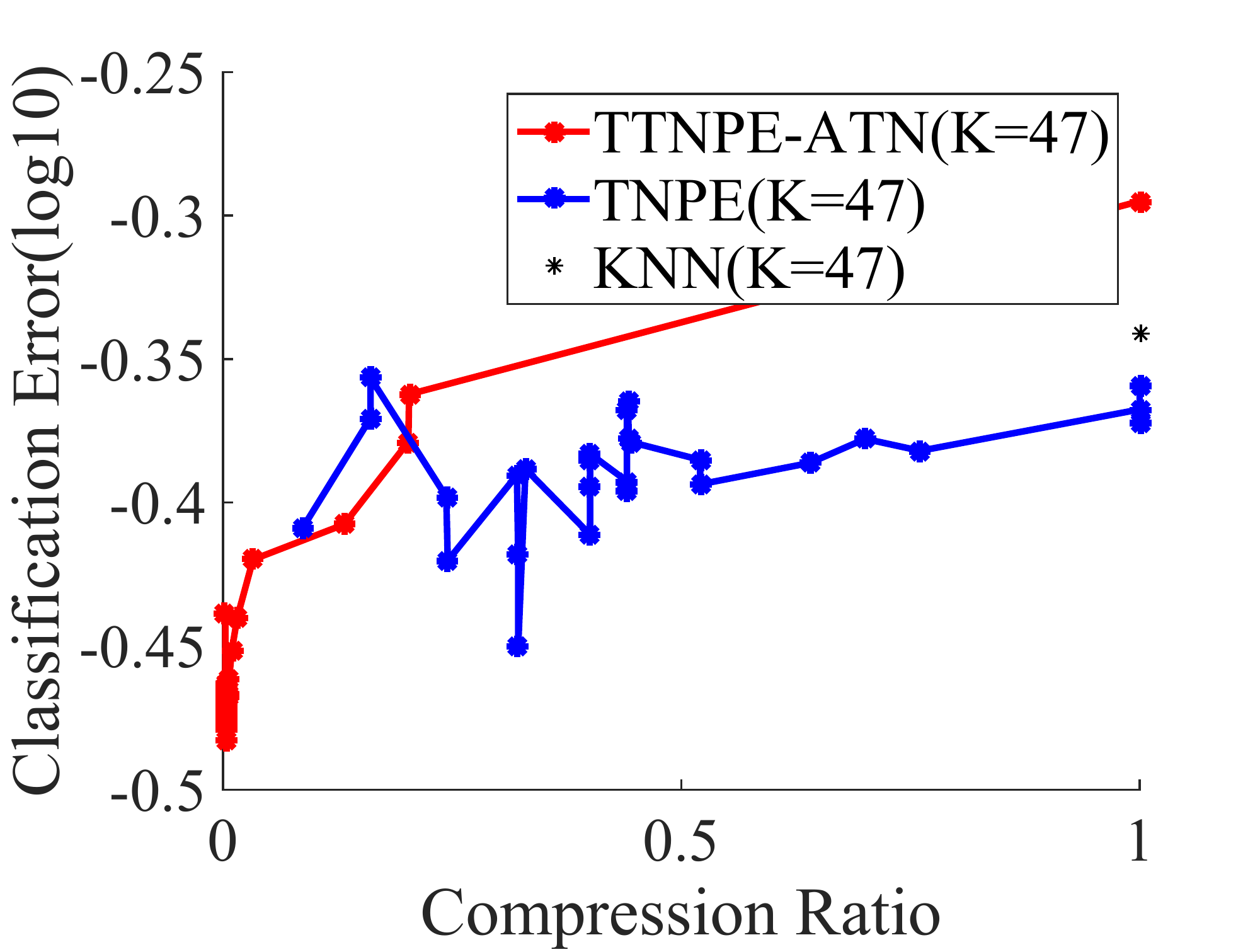}
\includegraphics [trim=.1in .2in .1in .1in, keepaspectratio, width=0.32\textwidth] {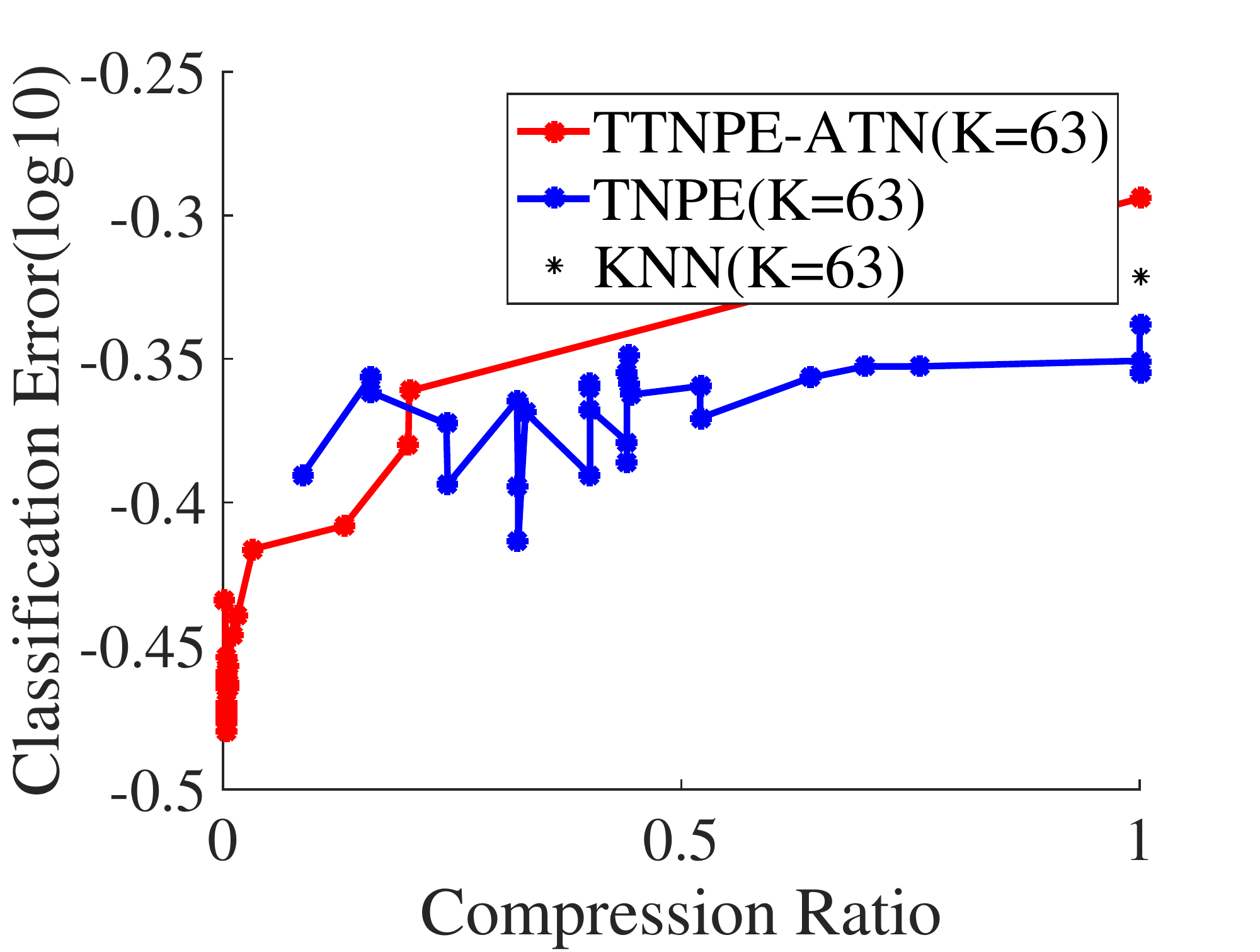}
\centering
\caption{\small Classification Error in $\log10$ scale for Finance for the three models.  }
\label{TTNPE_Finance}
\end{figure*}
In this section, tensor embedding method is applied to four year stock price data to determine whether the stock belongs to financial or technology sector. The stock prices used in this section are the daily adjusted closing prices for the top 400 companies, ranked by the market capital as of the end of 2017,  from financial  and technology sectors, respectively. 
The data is collected from 01/10/2014 to 12/29/2017 using  \cite{matfinance}, and the daily return of each stock is computed to be used as data. We did not use the absolute stock prices, but the return rates over these days to avoid the information in the absolute value of the stock price. The time-range mentioned above had 1001 business days, thus giving us 1000 data points for stock returns. 300 stocks from each sector (out of 400) are randomly sampled for training and the remaining data are used for testing.  Each time series is reshaped to a 3rd mode tensor $10\times10\times10$ for tensor embedding analysis. In the TTNPE-ATN, TNPE, and KNN algorithms, 31, 47, and 63 neighbors are selected for implementing the algorithm, respectively. Large number of neighbors empirically gives better and stable performance. Fig. \ref{TTNPE_Finance} illustrates the average results of 10 independent experiments over random choice of 300 training data  for each of the two sectors.


Financial data is known to be noisy. However, we note that  both the TNPE and TTNPE-ATN embedding algorithms outperform KNN,  thus the low dimensional tensor embedding is able to better reduce noise from the data. 
TTNPE-ATN algorithm classifies data more accurately in the low compression ratio regime while starts to degrade for  compression ratio greater than $0.1$, which is mainly due to over-fitting the noise. However, TTNPE-ATN still outperforms TNPE when the compression ratio is smaller than 0.3. 
}

\section{Conclusion}\label{concl}
This paper proposes a novel algorithm for non-linear Tensor Train Neighborhood Preserving Embedding (TTNPE-ATN) for tensor data classification. 
We investigate the tradeoffs between error, storage, and computation and evaluate the method on several vision datasets. 
We further show that TTNPE-ATN algorithm exhibits improved classification performance and better dimensionality reduction among the baseline approaches, and has lower computational complexity as compared to Tucker neighborhood preserving embedding method. 
In the future, we will investigate the convergence of tensor network optimization and provide the theoretical gap between TTNPE-ATN and TTNPE-TN. { While there has been work on parameter selection for matrix-based approaches \cite{yadav2015efficient,ubaru2016fast}, finding the  thresholding parameter for TTNPE is an interesting future research direction. }

\appendices
\begin{figure*}[htbp]
\centering
\includegraphics[trim=0in 0in 0in 0in, clip, width=.9\linewidth]{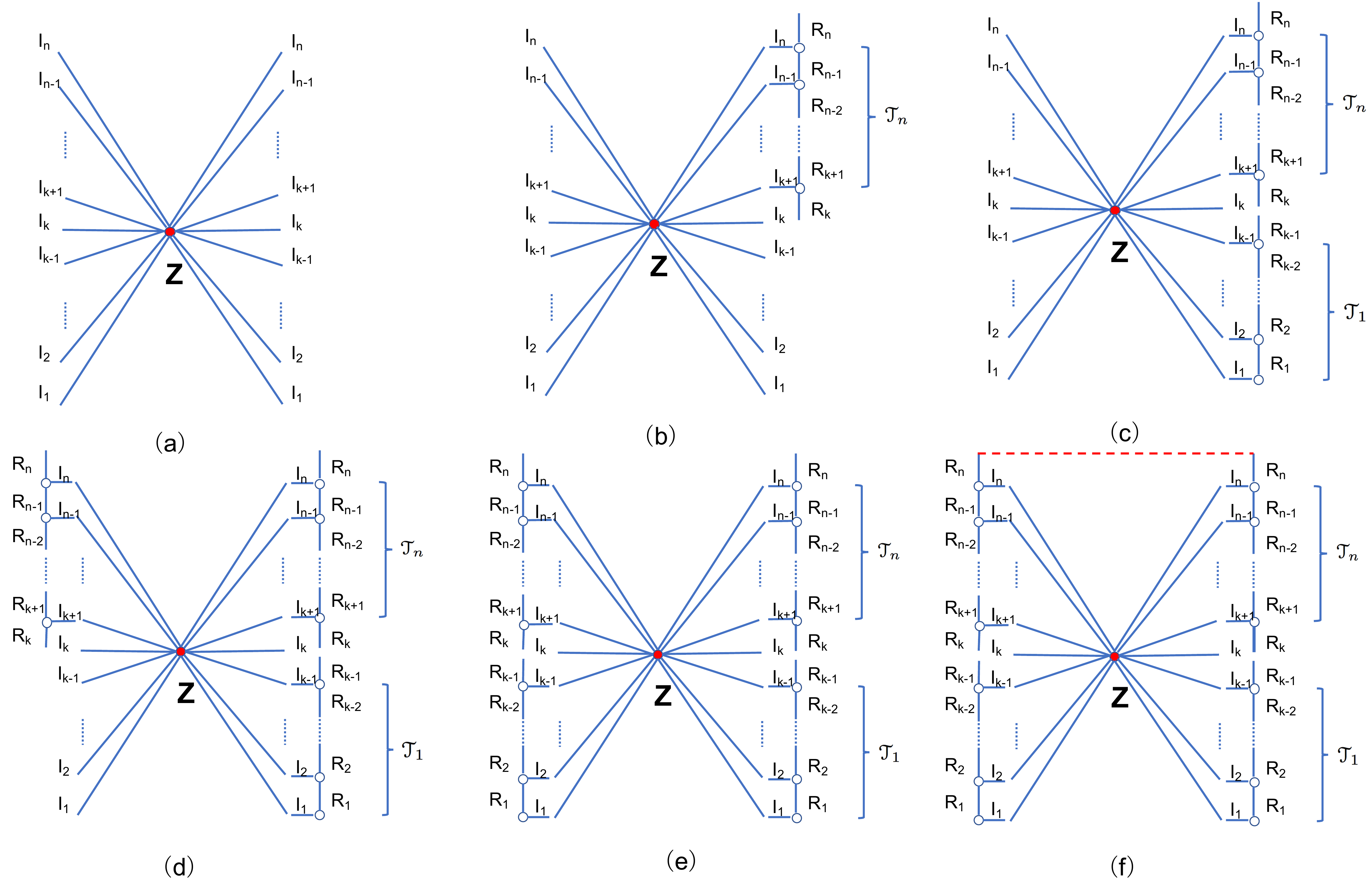}
\caption{\small Tensor network merging operation to compute  $\mathscr{A}$.
(a) Tensor $\mathscr{Z}$ 
(b) Tensor $\mathscr{A}_b$ 
(c) Tensor $\mathscr{A}_c$  
(d) Tensor $\mathscr{A}_d$ 
(e) Tensor $\mathscr{A}_e$ 
(f) Tensor $\mathscr{A}$ }
\label{TN_Merge}
\vspace{-.2in}
\end{figure*}

\begin{figure*}[htbp]
	\centering
	\includegraphics[trim=0in 0in 0in 0in, clip, width=.9\linewidth]{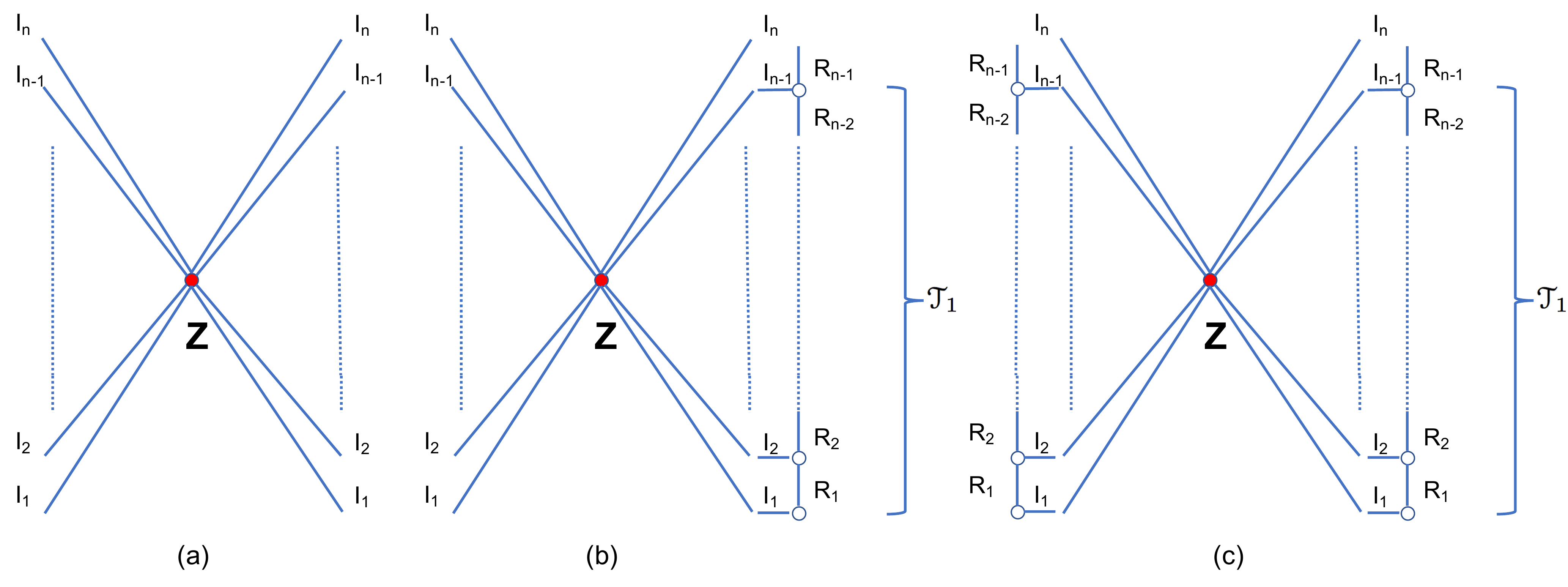}
	\caption{\small 
		Tensor network merging operation to compute  $\mathscr{B}$.
		(a) Tensor $\mathscr{Z}$ 
		(b) Tensor $\mathscr{B}_b$
		(c) Tensor $\mathscr{B}$}
	\label{TN_Merge2}
	\vspace{-.2in}
\end{figure*}

\section{Proof of Lemma \ref{equivalence}}\label{Lemma1}
The $(m,n)^\text{th}$ entry in the result gives 
\begin{equation}
\begin{split}
	&(\mathscr{A} \times^{1,\cdots,n}_{2,\cdots,n+1} \mathscr{B})_{m,n} \\
=  	&\sum_{r_1,\cdots, r_n} \mathscr{A}(m, r_1, \cdots, r_n) \mathscr{B}(r_1,\cdots, r_n, n),
\end{split}
\end{equation}
which is the same as the $(m,n)\text{th}$ entry given by ${\bf A} \times {\bf B}$.

\section{Proof of Lemma \ref{Lemma4}}\label{ProofLemma4}
Let ${\bf B}_j ={\bf L}(\mathscr{U}_1 \times_{3}^1 \cdots \times_{3}^1 \mathscr{U}_{j})$. { 
We first show ${\bf B}_{j+1} = ({\bf I}^{I_{j+1}} \otimes {\bf B}_j) \times {\bf L}(\mathscr{U}_{j+1})$. Using this, and induction (since the result holds for $j=1$), the result follows.
${\bf B}_j$ is a matrix of shape $(I_1I_2 \cdots I_j) \times R_j$. 
When $I_{j+1}=1$, $\mathscr{U}_{j+1}$ is a $3_\text{rd}$ order tensor of shape $R_{j} \times 1 \times R_{j+1}$, which is equivalent to a matrix of shape $R_{j} \times R_{j+1}$, thus ${\bf B}_{j+1} = {\bf B}_j \times \mathscr{U}_{j+1}$ becomes standard matrix multiplication. 
When $I_{j+1}>1$, the tensor  merging product is equivalent to the concatenation of $I_{j+1}$ matrix multiplications, which thus is  ${\bf B}_{j+1} = ({\bf I}^{I_{j+1}} \otimes {\bf B}_j) \times {\bf L}(\mathscr{U}_{j+1})$.
}

{
\section{Explanation of Tensor Network Merging Operation to compute $\mathscr{A}$ using \eqref{eq: target05}}\label{TNM}
Figure \ref{TN_Merge} shows the steps to compute $\mathscr{A}$. 
A tensor $\mathscr{Z}\in \mathbb{R}^{I_1\times\cdots\times I_n\times I_1\times\cdots\times I_n}$ in Fig \ref{TN_Merge} (a)  merged with tensor $\mathscr{T}_n \in \mathbb{R}^{R_k \times I_{k} \times \cdots \times I_n \times R_n}$ gives 
\begin{equation}
\mathscr{Z}  \times_{n+k+1, \cdots, 2n}^{2, \cdots, n-k+1} \mathscr{T}_n = \mathscr{A}_b \in \mathbb{R}^{I_1\times\cdots\times I_n\times I_1\times\cdots\times I_k \times R_k \times R_n},
\end{equation}
 as in Fig \ref{TN_Merge} (b), where the merged dimensions $I_{k+1}\times\cdots\times I_n$ are replaced by the non-merged dimension $R_k \times R_n$.
 Following the same logic, we have 
 \begin{equation}
 \mathscr{A}_b \times_{n+1,\cdots,n_k-1}^{1,\cdots, k-1} \mathscr{T}_1 = \mathscr{A}_c \in \mathbb{R}^{I_1\times \cdots \times I_n\times R_{k-1} \times I_k \times R_{k+1} \times R_n}\end{equation} 
 as in Fig \ref{TN_Merge} (c)), where the merged dimension $I_1 \times \cdots \times I_{k-1}$ are replaced by the non-merged dimension $R_{k-1}$. We further give the results to obtain tensor $\mathscr{A}_d$  and tensor $\mathscr{A}_e$ in Fig \ref{TN_Merge} (d) and (e) as follows
 \begin{equation}
 \begin{split}
  	&\mathscr{A}_c \times_{k+1\times \cdots \times n}^{2,\cdots, n-k+1} \mathscr{T}_n \\
  = 	&\mathscr{A}_d \in \mathbb{R}^{I_1 \times \cdots \times I_k \times R_k \times R_n \times R_{k-1} \times I_k \times R_{k+1} \times R_n}
 \end{split}
 \end{equation}
 and
  \begin{equation}
  \begin{split}
  	&\mathscr{A}_d \times_{1\times \cdots \times k-1}^{1\times \cdots \times k-1} \mathscr{T}_1 \\
  = 	&\mathscr{A}_e \in \mathbb{R}^{R_{k-1} \times I_k \times R_{k+1} \times R_n  \times R_{k-1} \times I_k \times R_{k+1} \times R_n}
  \end{split}
 \end{equation}
 The red marked trace operation in Fig. \ref{TN_Merge}(f) gets the trace along the $4^{th}$ and $8^{th}$ mode of $\mathscr{A}_e$, thus tensor $\mathscr{A}$ is obtained by $\mathscr{A} = \text{tr}_4^{8} \left(\mathscr{A}_e \right).$

\section{Explanation of Computing $\mathscr{B}$ using \eqref{eq:tn}}\label{TNM2}
Figure \ref{TN_Merge2} shows the steps to compute $\mathscr{B}$.  Computing $\mathscr{B}$ follows the same logic as computing $\mathscr{A}$, and is simpler since $\mathscr{T}_n$ does not involve in the computation. The step-by-step computation in Fig. \ref{TN_Merge2} (b) and (c) are as follows
\begin{equation}
\mathscr{Z}\times_{n+1,\cdots, 2n-1}^{1,\cdots, n-1} \mathscr{T}_1 = \mathscr{B}_b\in \mathbb{R}^{I_1\times\cdots\times I_n \times R_{n-1} \times I_n } ,
\end{equation}
and
\begin{equation}
\mathscr{B}_b\times_{1,\cdots, n-1}^{1,\cdots, n-1} \mathscr{T}_1 = \mathscr{B}\in \mathbb{R}^{R_{n-1} \times I_n \times R_{n-1} \times I_n } .
\end{equation}
}

\bibliographystyle{IEEEtran}

\bibliography{nips}

\end{document}